\documentclass[11pt,journal]{article}

\usepackage{algorithm}
\usepackage{listings}
\usepackage{epsfig}
\usepackage{amssymb,amsmath}
\usepackage{lineno}
\usepackage{multirow}

\usepackage{svg}
\usepackage{calc}
\usepackage{subfig}
\usepackage{booktabs}
\usepackage{url}
\usepackage{authblk}

\floatname{algorithm}{Algorithm}

\DeclareMathOperator*{\argmin}{argmin}

\floatstyle{ruled}
\newfloat{algorithm}{thp}{lop}
\floatname{algorithm}{Algorithm}

\usepackage{setspace}
\usepackage{xcolor}

\usepackage{geometry}
\begin{document}

\title{Towards Better Exploiting Convolutional Neural Networks for Remote Sensing Scene Classification}

\author{Keiller Nogueira$^1$, Ot\'avio A. B. Penatti$^2$, Jefersson A. dos Santos$^1$}
\affil{$^1$Departamento de Ci\^{e}ncia da Computa\c{c}\~{a}o, Universidade Federal de Minas Gerais (UFMG)\\
Belo Horizonte, MG, Brazil -- CEP 31270-901 \\
 \{keiller.nogueira,jefersson\}@dcc.ufmg.br\\
 $^2$Advanced Technologies Group, SAMSUNG Research Institute\\
   Campinas, SP, CEP 13097-160, Brazil\\
   o.penatti@samsung.com}
  
\maketitle

\begin{abstract}
We present an analysis of three possible strategies for exploiting the power of existing convolutional neural networks (ConvNets) in different scenarios from the ones they were trained: full training, fine tuning, and using ConvNets as feature extractors.
In many applications, especially including remote sensing, it is not feasible to fully design and train a new ConvNet, as this usually requires a considerable amount of labeled data and demands high computational costs. 
Therefore, it is important to understand how to obtain the best profit from existing ConvNets.
We perform experiments with six popular ConvNets using three remote sensing datasets. 
We also compare ConvNets in each strategy with existing descriptors and with state-of-the-art baselines. 
Results point that fine tuning tends to be the best performing strategy.
In fact, using the features from the fine-tuned ConvNet with linear SVM obtains the best results.
We also achieved state-of-the-art results for the three datasets used.

\end{abstract}


\section{Introduction}

Encoding 
discriminating features from visual data 
is one of the most important steps in almost any 
computer vision problem, including in the remote sensing domain.
Since manual extraction of these features is not practical in most cases, during years, substantial efforts have been dedicated to develop automatic and discriminating visual feature descriptors~\cite{kumar2014detailed}. 
In the early years, 
most of such descriptors were based on pre-defined algorithms independently of the underlying problem, like color histograms and correlograms~\cite{kumar2014detailed,dosSantos2010:VISAPP}.
Then, descriptors based on visual dictionaries, the so-called Bag of Visual Words (BoVW), attracted the attention and have become the state-of-the-art for many years in computer vision~\cite{tsai2012bag,ACSIJ110,SivicVideoGoogle2003,GemertVisuaWordAmbiguityPAMI2010,SandeEvaluatingTPAMI2010,BoureauMidLevelCVPR2010,PerronninFisherECCV2010}.
Although the aforementioned visual feature extraction techniques have been successfully applied in several domains~\cite{chen2010handbook}, due to the specificities of remotely sensed data, many of these classical techniques 
are not straightforwardly applicable in the remote sensing domain.
Indeed, encoding of spatial information in remote sensing images is still considered an open and challenging task~\cite{benediktsson2013advances}.
Thus, there is a huge demand for feature extraction algorithms able to effectively encode spectral and spatial information, since it is the key to generating discriminating models for remote sensing images.

Recently, 
deep learning 
has become the new state-of-the-art solution for visual recognition.
Given its success, deep learning has been intensively used in several distinct tasks of different domains~\cite{deeplearningbook,bengio2009learning}, including remote sensing~\cite{penattideep,nogueira2015improving,yue2015spectral}.
In remote sensing, the use of deep learning is growing very quickly, since it has a natural ability to effectively encode spectral and spatial information 
based mainly on the data itself.
Methods based on deep learning have obtained state-of-the-art results in many different remote sensing applications, such as image classification~\cite{nogueira2015improving}, oil spill~\cite{fingas2014review}, poverty mapping~\cite{xie2015transfer}, and urban planning~\cite{tayyebi2011urban}.

Deep learning~\cite{deeplearningbook,bengio2009learning}, a branch of machine learning that refers to multi-layered neural networks, is basically an architecture of interconnected layers that can learn features and classifiers at once,
i.e., a unique network may be able to learn features and classifiers (in different layers) and adjust the parameters, at running time, based on accuracy, giving more importance to one layer than another depending on the problem.
End-to-end feature learning (e.g., from image pixels to semantic labels) is the great advantage of deep learning when compared to previously state-of-the-art methods~\cite{lecun2015deep}, such as mid-level (BoVW) and global low-level color and texture descriptors.
Amongst all deep learning-based networks, a specific type, called Convolutional (Neural) Networks, ConvNets or CNNs~\cite{deeplearningbook, bengio2009learning}, is the most popular for learning visual features in computer vision applications, including remote sensing. 
This sort of network relies on the natural stationary property of an image, i.e., the statistics of one part of the image are the same as any other part and information extracted at one part of the image can also be employed to other parts.
Furthermore, deep ConvNets usually obtain different levels of abstraction for the data, ranging from local low-level information in the initial layers (e.g., corners and edges), to more semantic descriptors, mid-level information (e.g., object parts) in intermediate layers and high level information (e.g., whole objects) in the final layers.

The exploration of the potentials of deep ConvNets can be a complex task because of several challenges, such as:
(i) complex tuning, since convergence cannot be totally confirmed (given its highly non-convex property),
(ii) ``black box'' nature,
(iii) high computational burden,
(iv) proneness to overfitting, and
(v) empirical nature of model development, which is over parameterized.
However, through years, researches have developed strategies to explore the potential of deep ConvNets in different domains and scenarios.

In this work, we evaluate and analyze three possible strategies of exploiting ConvNets: (i) full-trained ConvNets, (ii) fine-tuned ConvNets, and (iii) pre-trained ConvNets used as feature extractors.
In the first strategy (i), a (new or existing) network is trained from scratch obtaining specific visual features for the dataset. 
This approach is preferable since it gives full control of the architecture and parameters, which tends to yield a more robust and efficient network. 
However, it requires a considerable amount of data~\cite{deeplearningbook,bengio2009learning}, since the convergence of network is pruned to overfitting (and small datasets tends to magnify this problem).
This drawback makes almost impracticable to fully design and train a network from scratch for most remote sensing problems, since large datasets in this domain are unusual given that training data may require high costs with travel and other logistics~\cite{Tuia2011:JSTSP,dosSantos2013:JSTARS}.
In an attempt to overcome the issue of few data to train the network, data augmentation\footnote{Data augmentation is a technique that artificially enlarges the training set with the addition of replicas of the training samples under certain types of transformations that preserve the class labels.} techniques~\cite{krizhevsky2012imagenet} can be employed. 
However, for small datasets even data augmentation is not enough to avoid overfitting.
The other two strategies (ii and iii) rely on the use of pre-trained ConvNets, i.e., we can employ networks that were trained on different data from the data of interest.
In fact, these strategies benefit from the property that initial layers of ConvNets tend to be generic filters, like edge or color blob detectors, which are less dependent on the final application and could be used in a myriad of tasks.
The second strategy (ii) uses a pre-trained ConvNet and performs fine-tuning of its parameters (filter weights) using the remote sensing data of interest.
Usually, in this case, the earlier layers are preserved, as they encode more generic features, and final layers are adjusted to encode specific features of the data of interest~\cite{hinton2006fast,larochelle2009exploring}.
The third strategy (iii) simply uses a pre-trained ConvNet as a feature extractor, by removing the last classification layer and considering its previous layer (or layers) as feature vector of the input data.

Thus, in this paper, we analyze the aforementioned strategies of exploiting deep learning in the remote sensing domain.
The objective is to elucidate and discuss some aspects of ConvNets which are necessary to take the most advantage of their benefits.
To the best of our knowledge, no other study in the literature realizes such analysis neither in remote sensing nor in other computer vision applications.
We carried out a systematic set of experiments comparing the results of the different strategies mentioned using~\textbf{six} existing ConvNets in three remote sensing datasets. 
In practice, our analysis is designed to address the following research questions: what is the strategy that better exploits the benefits of existing ConvNets for the remote sensing domain?
Therefore, we can summarize the contributions of our paper as: 
(i) analysis of the generalization power of ConveNets for other remote sensing datasets,
(ii) comparative analisys of ConvNets and sucessful low-level and mid-level feature descriptors,
(iii) evaluation and analysis of three strategies to exploit existing convnets in different scenarios, 
(iv) evaluation of ConvNets with state-of-the-art baselines.

Our results point that fine-tuning tends to be the best performing strategy.
In addition, we obtained state-of-the-art classification results for the tree remote sensing datasets used.




The remainder of this paper is organized as follows. 
Section~\ref{sec:rel_work} presents related work.
Some background concepts related to ConvNets are presented in Section~\ref{sec:background}.
Section~\ref{sec:methodology} presents, in details, all strategies evaluated in this work.
The experimental setup, datasets, ConvNets and the descriptors evaluated in this paper are presented in Section~\ref{sec:experiments}. 
In Section~\ref{sec:results}, we present and discuss the results obtained. 
Finally, Section~\ref{sec:conclusions} concludes the paper.

\section{Related Work} \label{sec:rel_work}

Considerable efforts have been dedicated to the development of suitable feature descriptors for remote sensing applications~\cite{kumar2014detailed}.
Although several of these visual descriptors have been proposed or successfully used for remote sensing image processing~\cite{Yang2008:ICIP,dosSantos2010:VISAPP,Bouchiha2013:SIVP},
there are many applications that demand more specific techniques mainly due to the presence of non-visible information provided by multiple spectral bands and the lack of adaptation in relation to aerial scenes (when compared to traditional sort of images).
Towards this goal, Convolutional Neural Networks, ConvNets or CNNs, which are the most popular deep learning approach for computer vision, involve machine learning in the process of obtaining the best visual features for a given problem.
They are based on an end-to-end learning process, from raw data (e.g., image pixels) to semantic labels, which is an important advantage in comparison with previous state-of-the-art methods~\cite{lecun2015deep}.
However, during the learning step, only parameters are learned, not the complete deep architecture, 
like the number and types of layers and how they are organized.
Specifically, deep ConvNets have several drawbacks, such as impossibility to confirm convergence, ``black box'' nature, high computational cost, proneness to overfitting and empirical nature of model development.
In an attempt to alleviate these effects, 
some strategies can be used to better exploit existing ConvNets, which are: 
(i) fine-tuned ConvNets, and (i) pre-trained ConvNets used as feature extractors.
Aside the strategy to fully training a network from scratch, these two approaches assemble the principal strategies when working with ConvNets.
Thus, the objective of this paper is to evaluate these three strategies to explore existing deep neural networks.
Hence, in this section, we focus on analyzing existing works that also exploit ConvNets similarly to what we do. 

Train a (new or existing) network from scratch is preferable since it tends to give specific visual features for the dataset.
Also, this strategy gives full control of the architecture and parameters, which tends to yield a more robust network. 
However, it requires a considerable amount of data~\cite{deeplearningbook,bengio2009learning}.
During the years, successful ConvNets were the ones trained in large amount of data, such as ImageNet dataset~\cite{deng2009imagenet}, which has been used to train several famous architectures~\cite{krizhevsky2012imagenet,simonyan2014very,szegedy2014going}.
AlexNet, proposed by Krizhevsky et al.~\cite{krizhevsky2012imagenet}, was the winner of ImageNet Large Scale Visual Recognition Challenge (ILSVRC)~\cite{deng2009imagenet} in 2012 and mainly responsible for the recent popularity of neural networks.
GoogLeNet, presented in~\cite{szegedy2014going}, is the ConvNet architecture that won the ILSVRC-2014 competition (classification and detection tracks) while VGG ConvNets, presented in~\cite{simonyan2014very}, won the localization and classification tracks of the same competition.

Although huge annotated remote sensing data are unusual, there are many works, usually using reasonable datasets (more than 2,000 images), that achieved promising results by proposing the full training of new ConvNets~\cite{nogueira2015improving,makantasis2015deep,yue2015spectral}. 
Nogueira et al.~\cite{nogueira2015improving} proposed and fully trained a new ConvNet architecture to classify images from aerial and multispectral images. 
Makantasis et al.~\cite{makantasis2015deep} classified hyperspectral images through a new ConvNet with only two convolution layers achieving state-of-the-art in four datasets.
In~\cite{yue2015spectral}, the authors proposed a hybrid method combining principal component analysis, ConvNets and logistic regression to classify hyperspectral image using both spectral and spatial features.
Moreover, a myriad of deep learning-based methods appeared exploiting its benefits in the remote sensing community~\cite{guan2015deep,chen2014deep,Firat2014:ICPR,Zhang2015:TGRS}.

In general, ConvNets have a peculiar property: they all tend to learn first-layer features that resemble either Gabor filters, edge detectors or color blobs.
Supported by this characteristic, another strategy to exploit ConvNets is to perform fine-tuning of its parameters using the new data.
Specifically, fine-tuning realizes adjustment in the parameters of a pre-trained network by resuming the training of the network from a current setting of parameters but considering a new dataset.
In~\cite{girshick2014rich}, the authors showed that fine-tuning a pre-trained ConvNet on the target data can significantly improve the performance. 
Specifically, they fine-tuned AlexNet~\cite{krizhevsky2012imagenet} and outperformed results for semantic segmentation.
Zhao et al.~\cite{chatfield2014return} fine-tuned a couple of networks outperforming state-of-the-art results in classification of traditional datasets.
Several works~\cite{xie2015transfer,yue2015spectral}, in the remote sensing community, also exploit the benefits of fine-tuning pre-trained ConvNets.
In~\cite{xie2015transfer}, the authors evaluated a full-trained ConvNet against a fine-tuned one to detect poverty using remote sensing data.
Yue et al.~\cite{yue2015spectral} used fine-tuning method to classify hyperspectral images.

Based on aforementioned characteristics, ConvNets can also be exploited as a feature extractor.
Specifically, these features, usually called deep features, are obtained by removing the last classification layer and considering the output of previous layer (or layers).
In some recent studies~\cite{penattideep,chatfield2014return,hu2015transferring}, ConvNets have shown to perform well even in datasets with different characteristics from the ones they were trained with, without performing any adjustment, using them as feature extractors only and using the features according to the application (e.g., classification, retrieval, etc). 
In~\cite{chatfield2014return}, the authors evaluated deep features, combined or not with other features, for classification of traditional images.
In remote sensing domains, Penatti et al.~\cite{penattideep} evaluated the use of different ConvNets as feature extractors, achieving state-of-the-art results in two remote sensing datasets, outperforming several well-known visual descriptors.
Hu et al.~\cite{hu2015transferring} extracted features of several pre-trained ConvNets to perform classification of high-resolution remote sensing imagery.

Concerning the evaluation of different practices to exploit deep neural networks, Jarrett et al.~\cite{jarrett2009best} analyzed the best architecture and training protocol to explore deep neural networks., including unsupervised and supervised ones. 
Also, Larochelle et al.~\cite{larochelle2009exploring} studied the best way to train a neural networks, including greedy layer-wise and unsupervised training.
Our paper differs from others in the literature since we evaluate different possible practices to exploit existing ConvNets.
To the best of our knowledge, there is no other study in the literature that performs such analysis neither in remote sensing nor in other computer vision applications.
Our paper also performs a set of experiments comparing the results of the different strategies mentioned using~\textbf{six} existing ConvNets in three remote sensing datasets.

\section{Background Concepts} \label{sec:background}

This section formally presents some background concepts of convolutional neural networks, or simply ConvNets, a specific type of deep learning method.
These networks are generally presented as systems of interconnected processing units (neurons) which can compute values from inputs leading to an output that may be used on further units.
These neurons work in agreement to solve a specific problem, learning by example, i.e., a network is created for a specific application, such as pattern recognition or data classification, through a learning process.
As introduced, ConvNets were initially proposed to work over images, since they try to take leverage from the natural stationary property of an image, i.e., information extracted in one part of the image can also be applied to another region.
Furthermore, ConvNets present several other advantages: (i) automatically learn local feature extractors, (ii) are invariant to small translations and distortions in the input pattern, and (iii) implement the principle of weight sharing which drastically reduces the number of free parameters and thus increases their generalization capacity.
Next, we present some concept employed in ConvNets.

\subsection{Processing Units} \label{subsec:units}

As introduced, artificial neurons are basically processing units that compute some operation over several input variables and, usually, have one output calculated through the activation function.
Typically, an artificial neuron has a weight vector $W = (w_1, w_2,\cdots, w_n)$, some input variables $X = (x_1, x_2,\cdots, x_n)$ and a threshold or bias $b$. Mathematically, vectors $w$ and $x$ have the same dimension, i.e., $w$ and $x$ are in $\Re^n$.
The full process of a neuron may be stated as in Equation~\ref{eq:activ_linear}.

\begin{eqnarray} \label{eq:activ_linear}
	z = f \left( \sum_i^{N} X_i * W_i + b \right)
\end{eqnarray}
where $z$, $x$, $w$ and $b$ represent output, input, weights and bias, respectively. $f(\cdot):\Re\rightarrow\Re$ denotes an activation function.


Conventionally, a nonlinear function is provided in $f(\cdot)$. 
There are a lot of alternatives for $f(\cdot)$, such as sigmoid, hyperbolic, and rectified linear function.
The latter function is currently the most used in the literature.
Neurons with this configuration have several advantages when compared to others:
(i) work better to avoid saturation during the learning process,
(ii) induce the sparsity in the hidden units, and
(iii) do not face gradient vanishing problem\footnote{The gradient vanishing problem occurs when the propagated errors become too small and the gradient calculated for the backpropagation step vanishes, making it impossible to update the weights of the layers and to achieve a good solution.} as with sigmoid and tanh function.
The processing unit that uses the rectifier as activation function is called Rectified Linear Unit (ReLU)~\cite{nair2010rectified}.
The first step of the activation function of a ReLU is presented in Equation~\ref{eq:activ_linear} while the second one is introduced in Equation~\ref{eq:relu_activ}.

\begin{eqnarray} \label{eq:relu_activ}
	a = \begin{cases}
		z, \mathit{if} z > 0 \\
		0, otherwise
	\end{cases} \qquad \Leftrightarrow \qquad a = f(z) = max(0,z)
\end{eqnarray}

The processing units are grouped into layers, which are stacked forming multilayer networks. 
These layers give the foundation to others, such as convolutional and fully-connected layers.

\subsection{Network Components} \label{subsec:NN_components}

Amongst the different types of layers, the convolutional one is the responsible for capturing the features from the images.
The first layers usually obtain low-level features (like edges, lines and corners) while the others get high-level features (like structures, objects and shapes).
The process made in this type of layer can be decomposed into two phases:
(i) the convolution step, where a fixed-size window runs over the image, with some stride~\footnote{Stride is the distance between the centers of each window considering two steps.}, defining a region of interest, and
(ii) the processing step, that uses the pixels inside each window as input for the neurons that, finally, perform the feature extraction from the region. 
Formally, in the latter step, each pixel is multiplied by its respective weight generating the output of the neuron, just like Equation~\ref{eq:activ_linear}.
Thus, only one output is generated concerning each region defined by the window.
This iterative process results in a new image (or feature map), generally smaller than the original one, with the visual features extracted.
Many of these features are very similar, since each window may have common pixels, generating redundant information. 
Typically, after each convolutional layer, there are pooling layers that were created in order to reduce the variance of features by computing some operation of a particular feature over a region of the image. Specifically, a fixed-size window runs over the features extracted by the convolutional layer and, at each step, a operation is realized to minimize the amount and optimize the gain of the features. 
Two operations may be realized on the pooling layers: the max or average operation, which selects the maximum or mean value over the feature region, respectively. This process ensures that the same result can be obtained, even when image features have small translations or rotations, being very important for object classification and detection.
Thus, the pooling layer is responsible for sampling the output of the convolutional one preserving the spatial location of the image, as well as selecting the most useful features for the next layers.

After several convolutional and pooling layers, there are the fully-connected ones, which take all neurons in the previous layer and connect them to every single neuron in its layer.
The previous layers can be convolutional, pooling or fully-connected, however the next ones must be fully-connected until the classifier layer, because the spatial notion of the image is lost in this layer.
Since a fully-connected layer occupies most of the parameters, overfitting can easily happen.
To prevent this, the dropout method~\cite{srivastava2014dropout} was employed.
This technique randomly drops several neuron outputs, which do not contribute to the forward pass and backpropagation anymore.
This neuron drops are equivalent to decreasing the number of neurons of the network, improving the speed of training and making model combination practical, even for deep  networks. Although this method creates networks with different architectures, those networks share the same weights, permitting model combination and allowing that only one network is needed at test time.

Finally, after all convolution, pooling and fully-connected layers, a classifier layer may be used to calculate the class probability of each instance. The most common classifier layer is the softmax one~\cite{bengio2009learning}, based on the namesake function.
The softmax function, or normalized exponential, is a generalization of the multinomial logistic function that generates a K-dimensional vector of real values in the range $(0, 1)$ which represents a categorical probability distribution.
Equation~\ref{eq:softmax_func} shows how softmax function predicts the probability for the $j$th class given a sample vector $X$.

\begin{eqnarray} \label{eq:softmax_func}
	h_{W,b}(X) = P(y=j|X; W, b) = \frac{\exp^{X^T W_j}}{\sum^{K}_{k=1} \exp^{X^T W_k}}
\end{eqnarray}
where $j$ is the current class being evaluated, $X$ is the input vector, and $W$ represent the weights.

In addition to all these processing layers, there are also normalization ones, such as Local Response Normalization (LRN)~\cite{krizhevsky2012imagenet} layer. This is the most useful when using processing units with unbounded activations (such as ReLU), because it permits the local detection of high-frequency features with a big neuron response, while damping responses that are uniformly large in a local neighborhood.

\subsection{Training} \label{subsec:train}

After modeling a network, in order to allow the evaluation and improvement of its results, a loss function needs to be defined, even because the goal of the training is to minimize the error of this function, based on the weights and bias, as presented in Equation~\ref{eq:argmin}.
Amongst several functions, the log loss one has become more pervasive because of exciting results achieved in some problems~\cite{krizhevsky2012imagenet}.
Equation~\ref{eq:log_loss} presents a general log loss function, without any regularization (or weight decay) term, which is used to prevent overfitting.

\begin{eqnarray} \label{eq:argmin}
	\argmin_{W,b}[\mathcal{J}(W,b)]
\end{eqnarray}

\begin{equation} \label{eq:log_loss}
	\begin{split}
		\mathcal{J}(W,b) = - \frac{1}{N} \sum^{N}_{i=1} ( y^{(i)} \times \log h_{W,b}(x^{(i)}) + \\
		(1-y^{(i)}) \times \log (1-h_{W,b}(x^{(i)})))
	\end{split}
\end{equation}

where $y$ represents a possible class, $x$ is the data of an instance, $W$ the weights, $i$ is an specific instance, and $N$ represents the total number of instances.

With the cost function defined, the ConvNet can be trained in order to minimize the loss by using some optimization algorithm, such as Stochastic Gradient Descent (SGD), to gradually update the weights and bias in search of the optimal solution:

\begin{equation*} \label{eq:update_weights}
	\begin{split}
		W_{ij}^{(l)} = W_{ij}^{(l)} - \alpha \frac{\partial\mathcal{J}(W,b)}{\partial W_{ij}^{(l)}} \\
		b_{i}^{(l)} = b_{i}^{(l)} - \alpha \frac{\partial\mathcal{J}(W,b)}{\partial b_{i}^{(l)}}
	\end{split}
\end{equation*}
where $\alpha$ denotes the learning rate, a parameter that determines how much an updating step influences the current value of the weights, i.e., how much the model learns in each step.

However, as presented, the partial derivatives of the cost function, for the weights and bias, are needed.
To obtain these derivatives, the backpropagation algorithm is used.
Specifically, it must calculate how the error changes as each weight is increased or decreased slightly.
The algorithm computes each error derivative by first computing the rate at which the error $\delta$ changes as the activity level of a unit is changed.
For classifier layers, this error is calculated considering the predicted and desired output.
For other layers, this error is propagated by considering the weights between each pair of layers and the error generated in the most advanced layer.

The training step of a ConvNet occurs in two steps:
(i) the feed-forward one, that passes the information through all the network layers, from the first until the classifier one, usually with high batch size~\footnote{Batch size is a parameter that determines the number of images that goes through the network before the weights and bias are updated.}, and
(ii) the backpropagation one, which calculates the error $\delta$ generated by the ConvNet and propagates this error through all the layers, from the classifier until the first one. As presented, this step also uses the errors to calculate the partial derivatives of each layers for the weights and bias.

\section{Strategies for Exploiting ConvNets}  \label{sec:methodology}

This section aims at explaining the most common strategies of employing existing ConvNets in different scenarios from the ones they were trained for. 
As introduced, training a deep network from scratch requires a considerable amount of data as well as a lot of computational power.
In many problems, few labeled data is available, therefore training a new network is a challenging task.
Hence, it is common to use a pre-trained network either as a fixed feature extractor for the task of interest or as an initialization for fine-tuning the parameters.
We describe these strategies, including their advantages and disadvantages, in the next subsections.
Section~\ref{fulltrained} describes about the full training of a new network while 
Section~\ref{finetuned} presents the fine-tuning process. 
Finally, Section~\ref{pretrained} explains the use of deep ConvNets as a fixed feature extractor.




\subsection{Full-trained Network} \label{fulltrained}


The strategy to train a network from scratch (with random initialization of the filter weights) is the first one to be thought when training ConvNets.  
This process is useful when the dataset is large enough to make a network converge and has several advantages, such as:
(i) extractors tuned specifically for the dataset, which tend to generate more accurate features, and
(ii) full control of the network.
However, fully training a network from scratch requires a lot of computational and data resources~\cite{deeplearningbook,bengio2009learning}, since the convergence of network is pruned to overfitting. 
Also, the convergence of a ConvNet can not be totally confirmed (given its highly non-convex property) making tuning not so trivial.

There are basically two options of training a deep network that we can fit in the case of full training.
The first one is by fully designing and training a new architecture, including
number of layers, neurons and type of activations, the number of iterations, weight decay, learning rate, etc.
The other option is by using an existing architecture and fully training its weights to the target dataset.
In this last case, the architecture and parameters (weight decay, learning rate, etc) are not modified at all.

\begin{figure}[t!]
	\centering
	\includegraphics[width=0.8\textwidth]{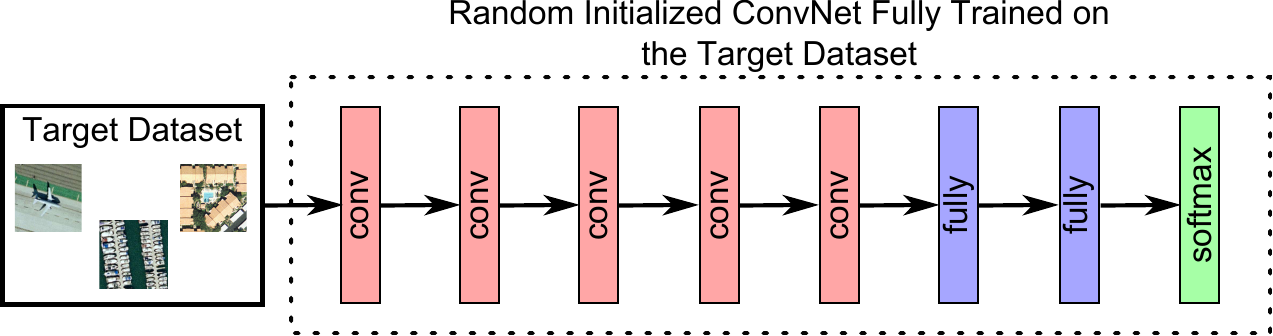}
	\caption{Overview of a ConvNet being fully trained.
		Weights from the whole network are randomly initialized and then trained for the target dataset.}
	\label{fig:fullytrained}
\end{figure}

Full training is expected to achieve better results when compared to other strategies, since the deep network learns specific features for the dataset of interest.
In this paper, we do not evaluate the option of creating a whole new ConvNet, because of challenges presented above.
Instead, we consider the full training of existing architectures in new datasets.
This strategy is illustrated in Figure~\ref{fig:fullytrained}.

\subsection{Fine-tuned Network} \label{finetuned}

When the new dataset is reasonable large, but not enough to full train a new network, fine-tuning is a good option to extract the maximum effectiveness from pre-trained deep ConvNets, since it can significantly improve the performance of the final classifier.
Fine-tuning is based on a curious property of modern deep neural network: they all tend to learn first-layer features that resemble either Gabor filters, edge or color blob detectors, independently of the training data.
More specifically, earlier layers of a network contain generic features that should be useful to many tasks, but later layers become progressively more specific to the details of the classes contained in the original dataset (i.e., the dataset in which the deep ConvNet was originally trained).
Supported by this property, the initial layers can be preserved while the final ones should be adjusted to suit the dataset of interest.

Fine-tuning performs a fine adjustment in the parameters of a pre-trained network by resuming the training of the network from a current setting of parameters but considering a new dataset, aiming at accuracy improvements. 
In other words, fine-tuning uses the parameters learned from a previous training of the network on a specific dataset and, then, adjusts the parameters from the current state for the new dataset, improving the performance of the final classifier.

Based on aforementioned characteristics, there are two possible approaches of performing fine-tuning in a pre-trained network, both exploited in this work:
(i) fine-tune all layers, and
(ii) fine-tune only higher-level layers keeping some of the earlier layers fixed (due to overfitting concerns).
It is important to emphasize that in both scenarios, the search space is bounded to just small variations in each step, since the learning rate is initialized with reduced value.
Specifically, in the first case, some layers (usually the final ones, such as the classification layer, since the number of classes tend to be different) have weights ignored, being randomly initialized. 
These layers have the learning rate increased, so they can learn faster and converge, while the other layers may also change weights by very small variations, since they use the reduced value of the learning rate without any augmentation.
By doing this, the first layers can use the information previously learned with just few adjustments to the dataset of interest, and at the same time, the final layers can really learn based only on the new dataset.
In the second case, the initial layers are frozen to keep the generic features already learned, while the final layers are adjusted using the increased value of the learning rate.

\begin{figure}[t!]
	\centering
	\includegraphics[width=0.8\textwidth]{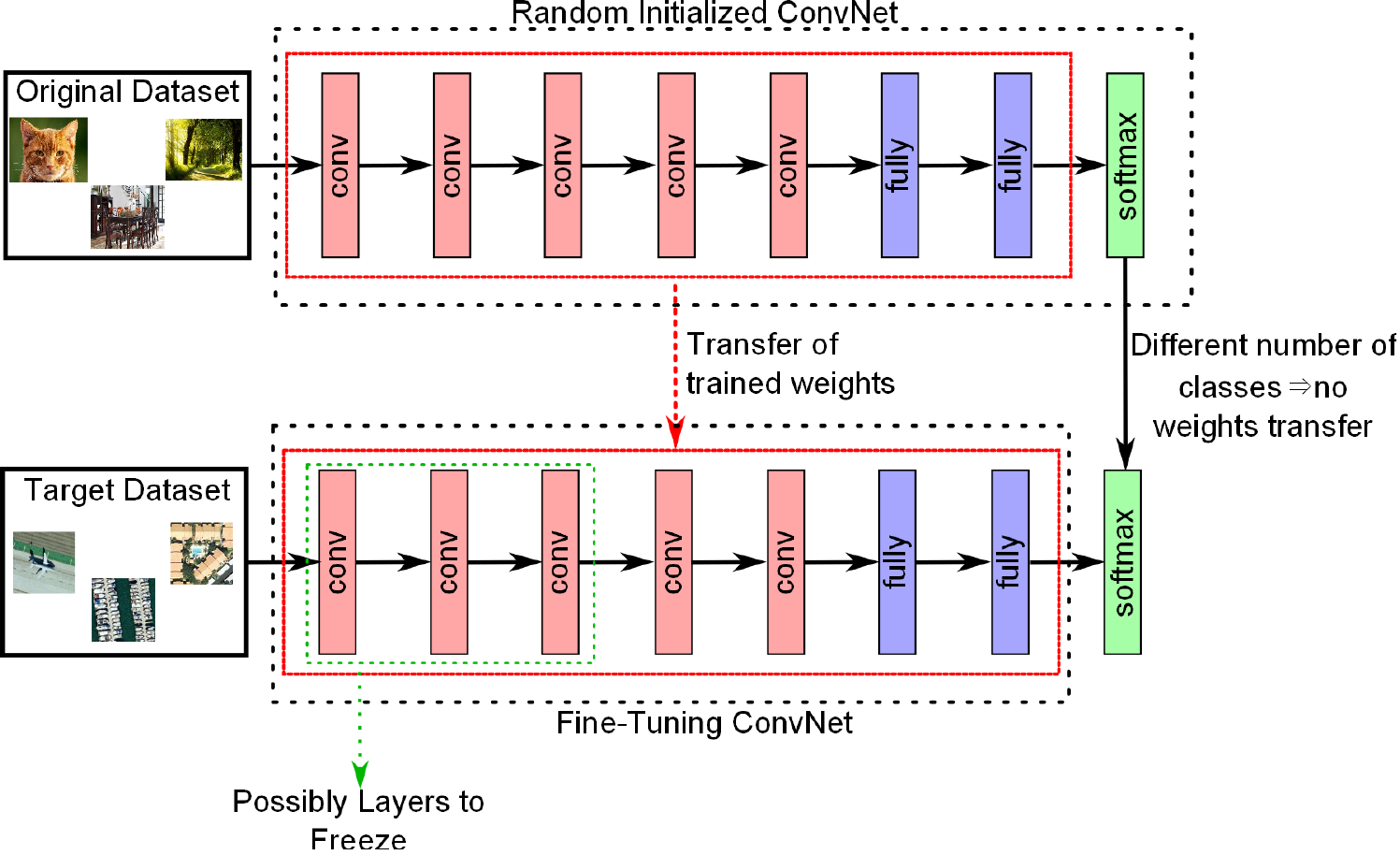}
	\caption{Overview of two options for the fine-tuning process. In one of them, all layers are fine-tuned according to the target dataset, but final layers have increased learning rates. In the other option, weights of initial layers can be frozen and only final layers are tuned.}
	\label{fig:finetuning}
\end{figure}

These two options of fine tuning are illustrated in Figure~\ref{fig:finetuning}.

\subsection{ConvNet as a Feature Extractor} \label{pretrained}

A pre-trained network can be used as a feature extractor for any image, since the generic features (learned in earlier layers) are less dependent on the final application and could be used in a myriad of tasks.
Specifically, features (usually, called deep features) can be extracted from any layer of a pre-trained network and then used in a given task.
Deep features trained on ImageNet (a dataset of everyday objects) have already shown remarkable results in applications like flower categorization~\cite{sunderhauf2014fine}, human attribute detection~\cite{hara2014fashion}, bird sub-categorization~\cite{ge2015fine}, scene retrieval~\cite{babenko2014neural}, and many others~\cite{chatfield2014return,girshick2014rich}, including remote sensing~\cite{penattideep,hu2015transferring}.
Furthermore, Razavian et al.~\cite{OverFeat2014} suggest that features obtained from deep learning should be the primary candidate in most visual recognition tasks.

The strategy of using pre-trained ConvNets as feature extractors is very useful given its simplicity, since no retraining or tuning is necessary. 
Moreover, one only needs to select the layer to be used, extract the deep features and use them combined with some machine learning technique, in case of a classification setup.
According to previous works~\cite{penattideep, OverFeat2014, chatfield2014return}, deep features can be extracted from the last layer before the classification layer (usually, a fully-connected one) and, then, used to train a linear classifier, 
which is the strategy employed in this paper.

\begin{figure}[t!]
	\centering
	\includegraphics[width=0.8\textwidth]{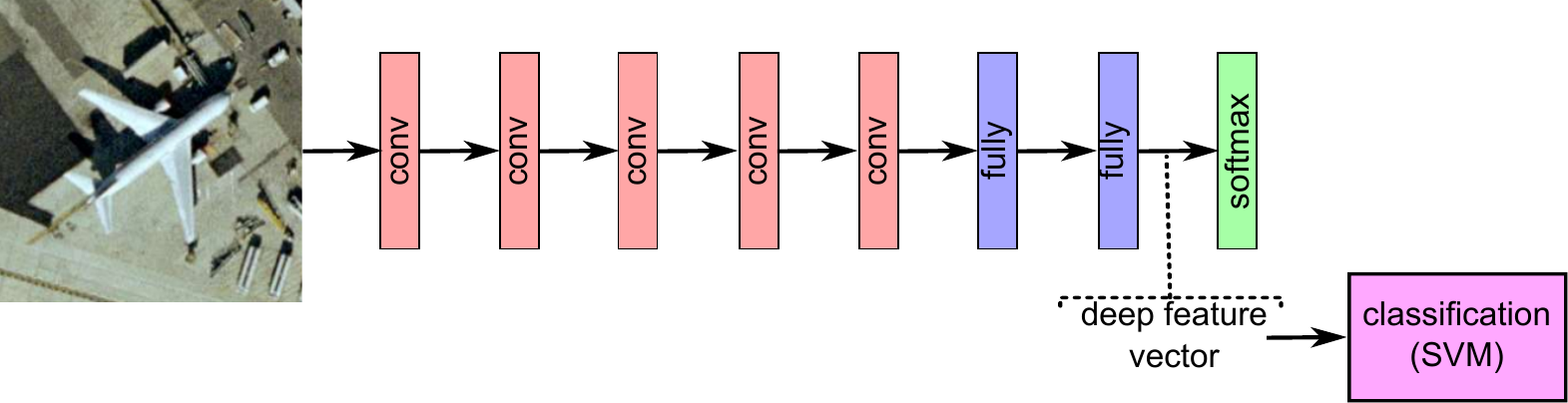}
	\caption{Overview of the use a ConvNet as feature extractor. The final classification layer is ignored and only the layer used to extract the deep features need to be defined.
		The figure shows the use of the features from the last layer before the classification layer, which is commonly used in the literature.}
	\label{fig:extraction}
\end{figure}

Figure~\ref{fig:extraction} illustrates how to use an existing ConvNet as a feature extractor.

\section{Experimental Setup} \label{sec:experiments}

The main objective of this paper is to evaluate which are the most suitable strategies to better exploit the power of deep features for remote sensing image scenes classification.
%
The details about the experiments conducted are presented in the following subsections.
The datasets are presented in Section~\ref{subsec:datasets}.
Section~\ref{subsec:descriptors} describes the low-level (global) and mid-level (BoVW) descriptors used while Section~\ref{subsec:cnn} presents the evaluated ConvNets.
Finally, Section~\ref{subsec:protocol} presents the protocol used in the experiments.

\subsection{Datasets} \label{subsec:datasets}

We have chosen remote sensing datasets with different visual properties in order to better evaluate the robustness and effectiveness of each strategy.
The first one, presented in Section~\ref{subsubsec:ucmerced_dataset}, is a multi-class land-use dataset that contains aerial high resolution scenes in the visible spectrum. 
The second one, presented in Section~\ref{subsubsec:rs19}, is a multi-class high resolution dataset with images collected from different regions all around the world.
The last one, presented in Section~\ref{subsubsec:coffee_dataset}, has multispectral high-resolution scenes of coffee crops and non-coffee areas.

\subsubsection{UCMerced Land-use}  \label{subsubsec:ucmerced_dataset}

This manually labelled and publicly available dataset~\cite{Yang2010:SIGSPATIAL} is composed of 2,100 aerial scene images with $256\times256$ pixels equally divided into 21 land-use classes selected from the United States Geological Survey (USGS) National Map.  
Thus, these images were obtained from different US locations, providing diversity to the dataset.
The 21 categories are: agricultural, airplane, baseball diamond, beach, buildings, chaparral, dense residential, forest, freeway, golf course, harbor, intersection, medium density residential, mobile home park, overpass, parking lot, river, runway, sparse residential, storage tanks, and tennis courts. 
Some class samples are shown in Figure~\ref{fig:merced_dataset}. 
It is remarkable the overlapping of some classes, such as ``dense residential'', ``medium residential'' and ``sparse residential'', which mainly differ in the density of structures.

\newcommand{\exFigSize}{0.12}

\begin{figure}[t!]
	\centering
	\scriptsize
	\subfloat[Dense Residential]{
		\includegraphics[width=\exFigSize\textwidth, keepaspectratio=true]{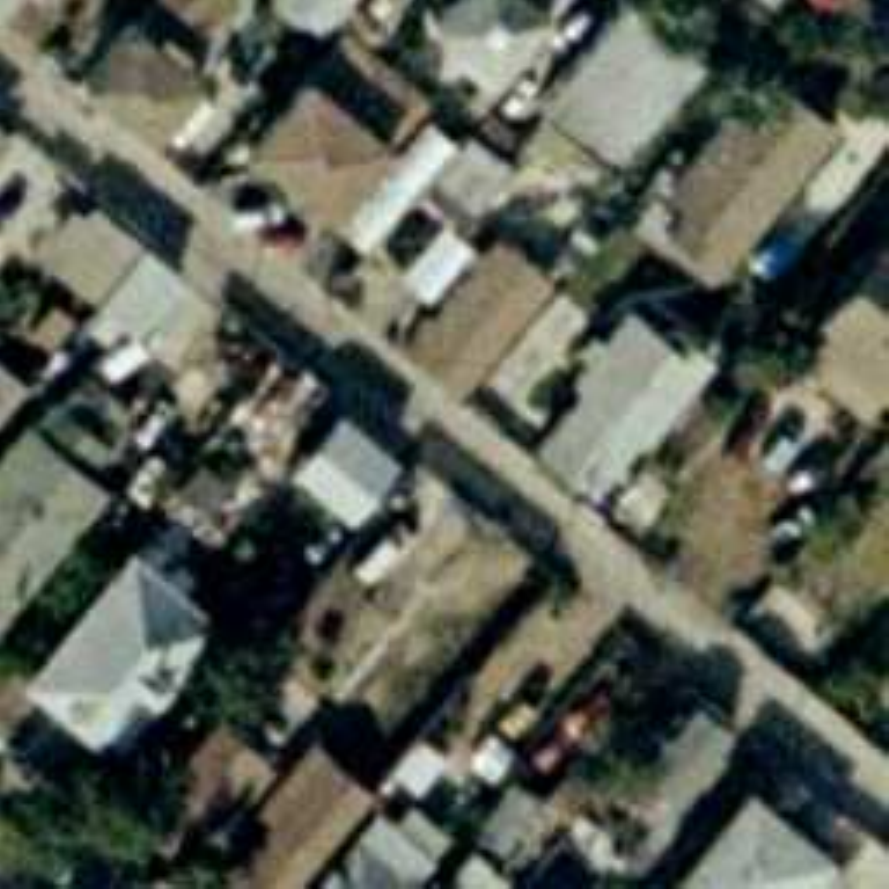}
		\includegraphics[width=\exFigSize\textwidth, keepaspectratio=true]{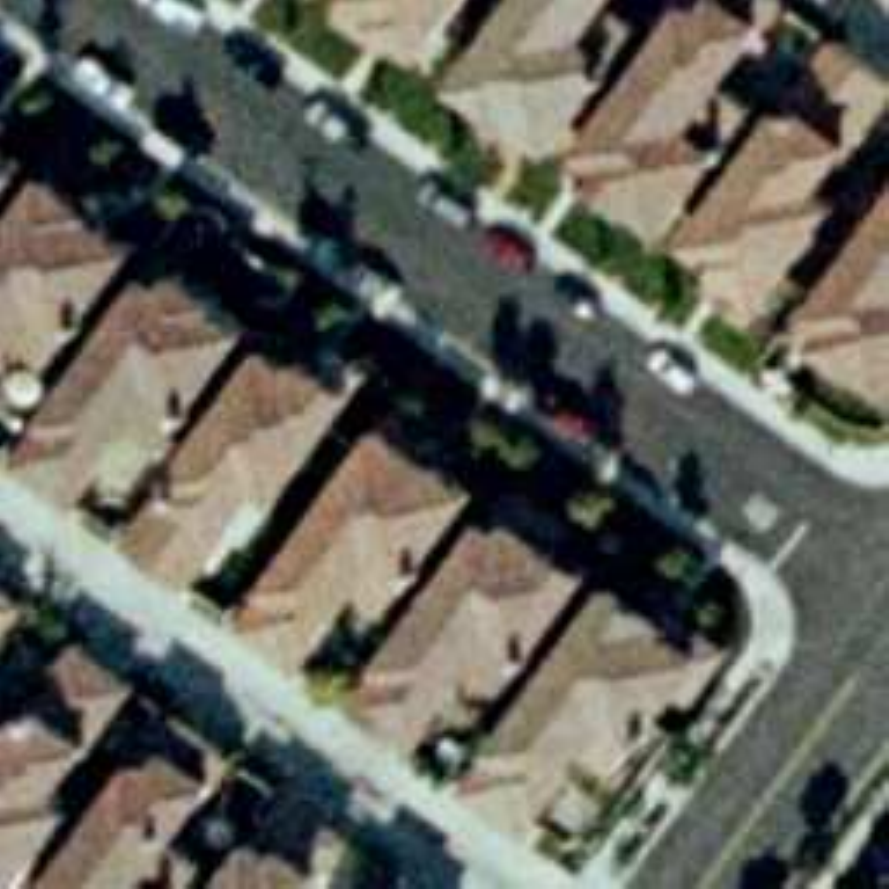}
	}
	\hspace{1mm}
	\subfloat[Medium Residential]{
		\includegraphics[width=\exFigSize\textwidth, keepaspectratio=true]{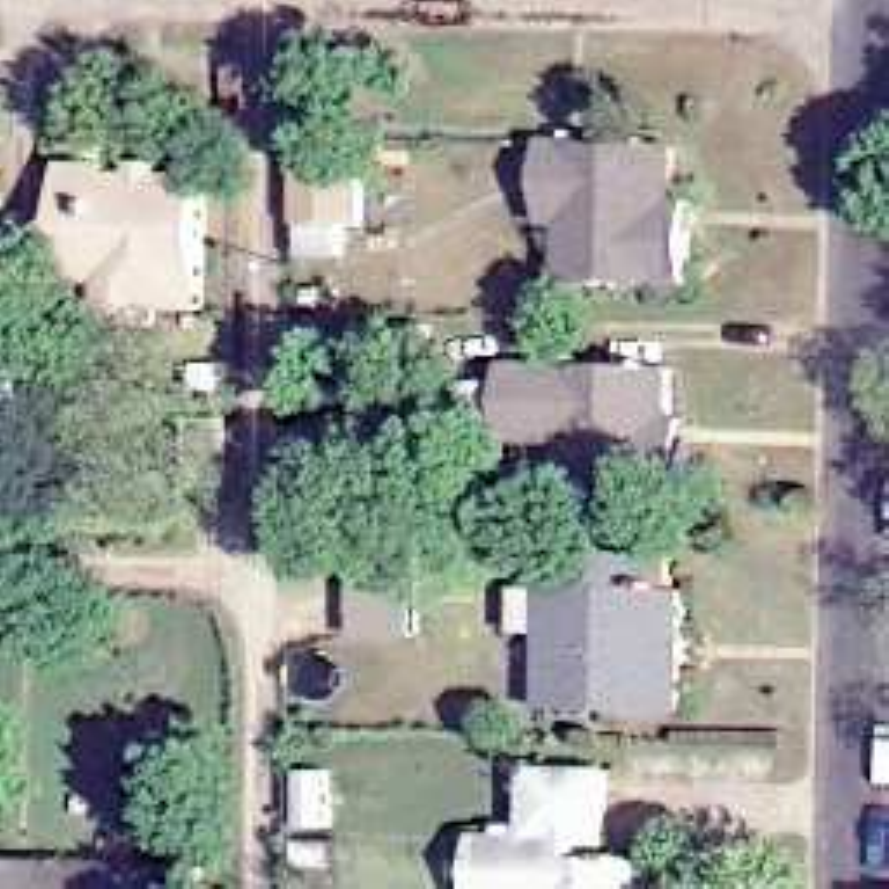}
		\includegraphics[width=\exFigSize\textwidth, keepaspectratio=true]{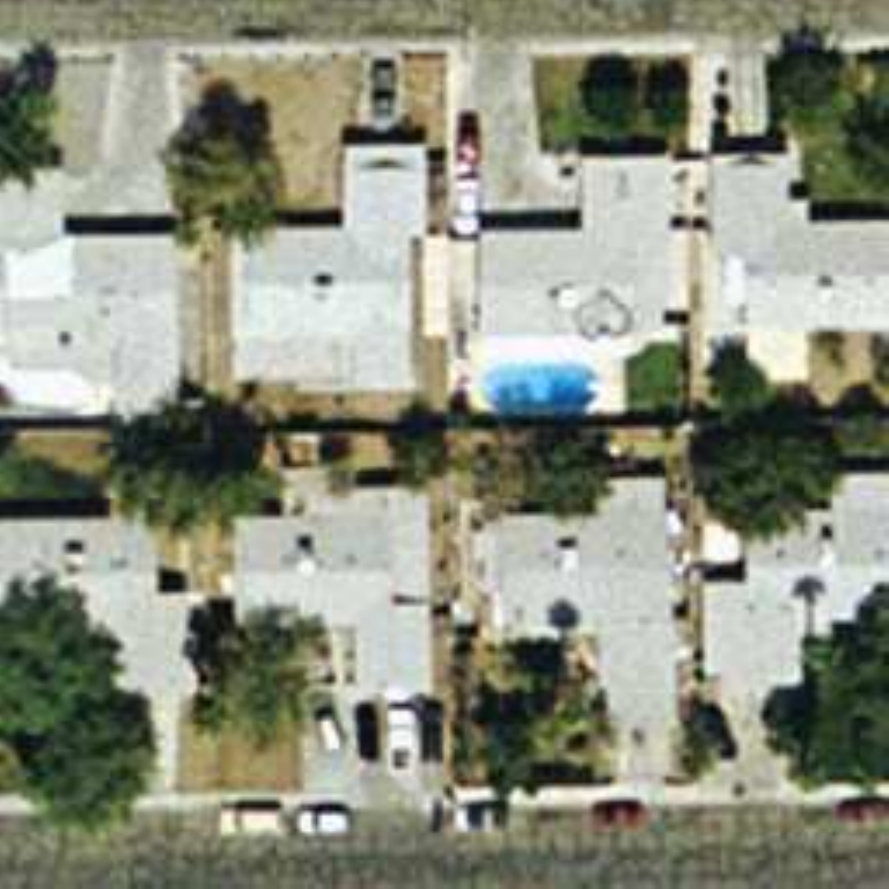}
	}
	\hspace{1mm}
	\subfloat[Harbor]{
		\includegraphics[width=\exFigSize\textwidth, keepaspectratio=true]{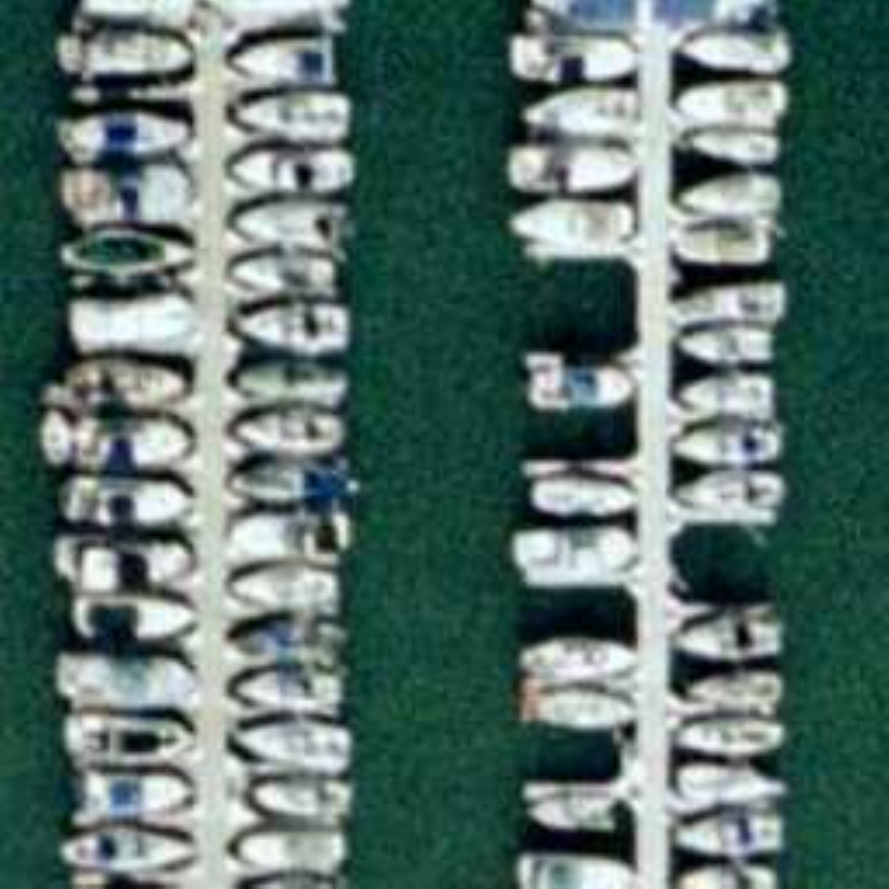}
		\includegraphics[width=\exFigSize\textwidth, keepaspectratio=true]{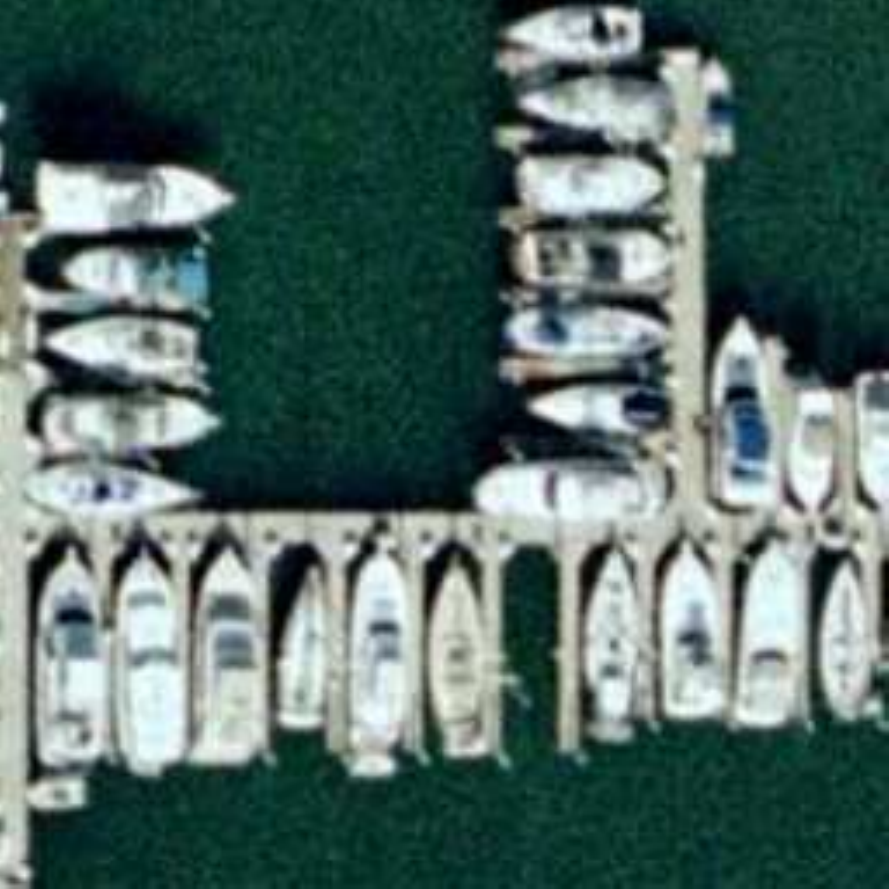}
	}
	\hspace{1mm}
	\subfloat[Freeway]{
		\includegraphics[width=\exFigSize\textwidth, keepaspectratio=true]{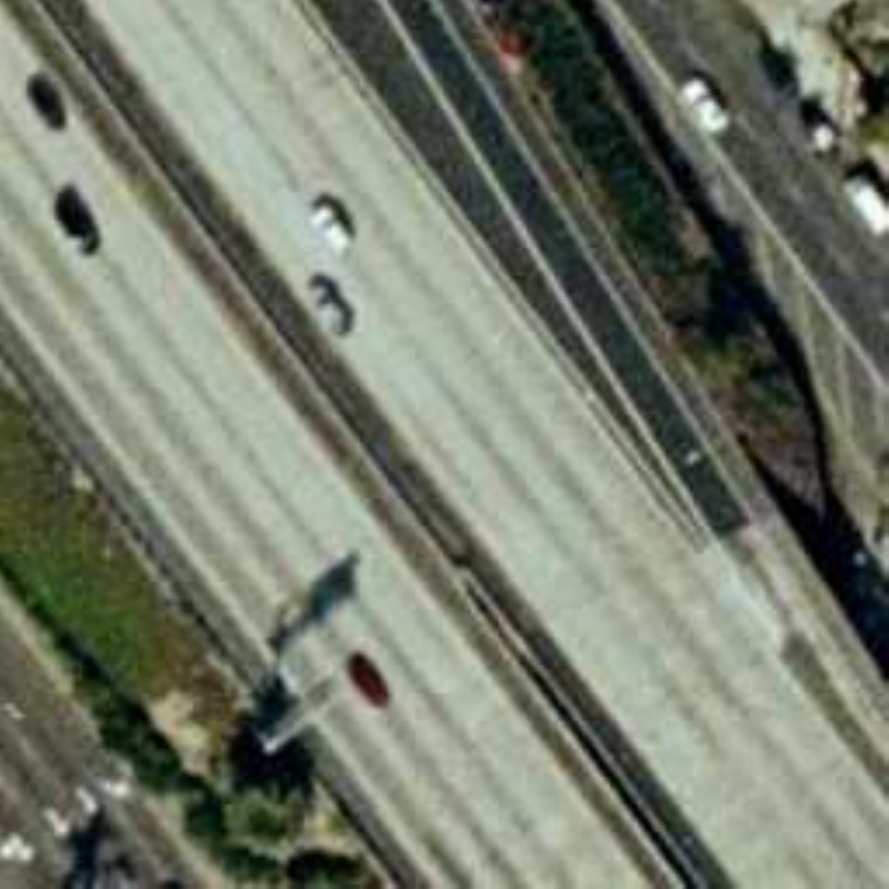}
		\includegraphics[width=\exFigSize\textwidth, keepaspectratio=true]{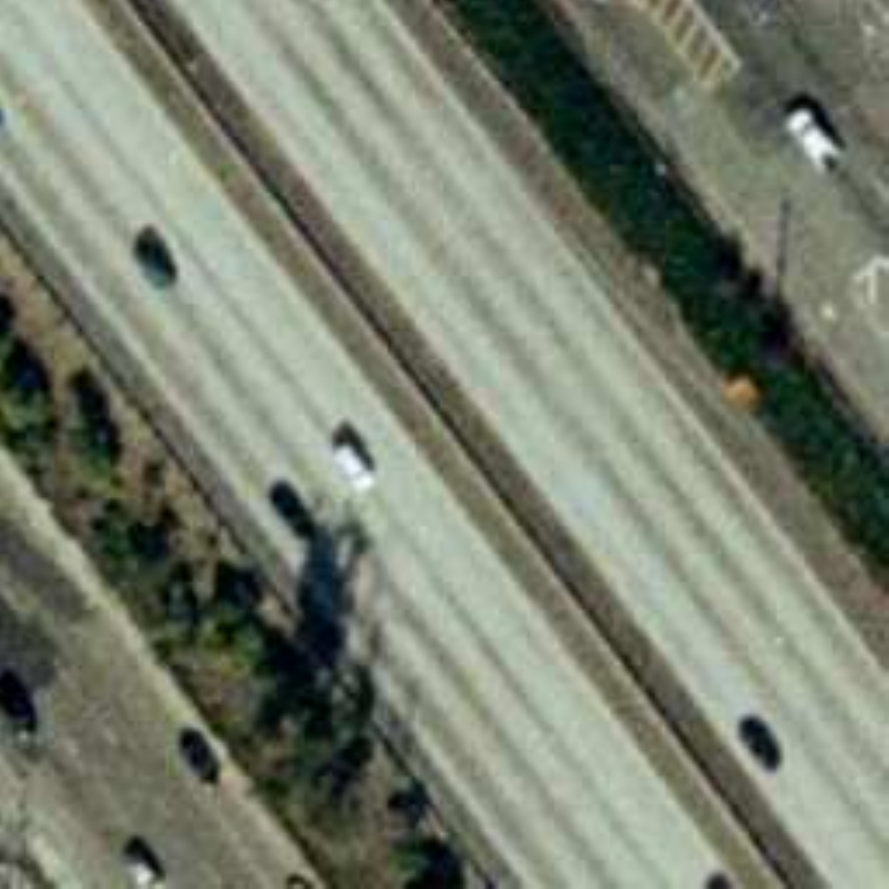}
	}
	\hspace{1mm}
	\subfloat[Runway]{
		\includegraphics[width=\exFigSize\textwidth, keepaspectratio=true]{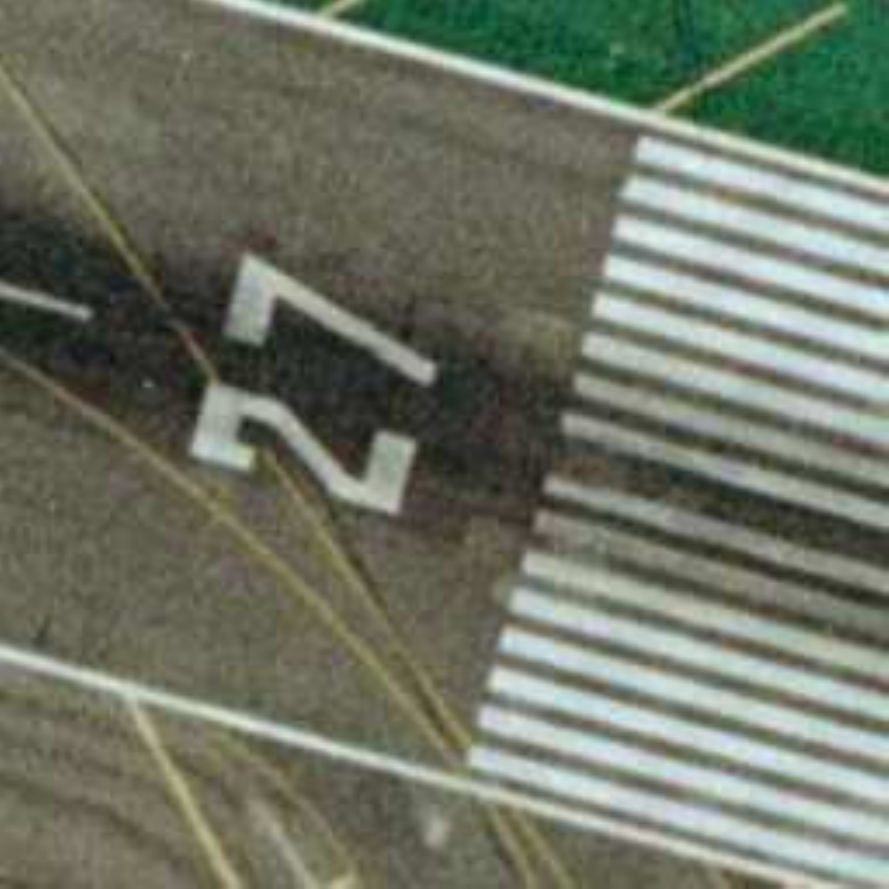}
		\includegraphics[width=\exFigSize\textwidth, keepaspectratio=true]{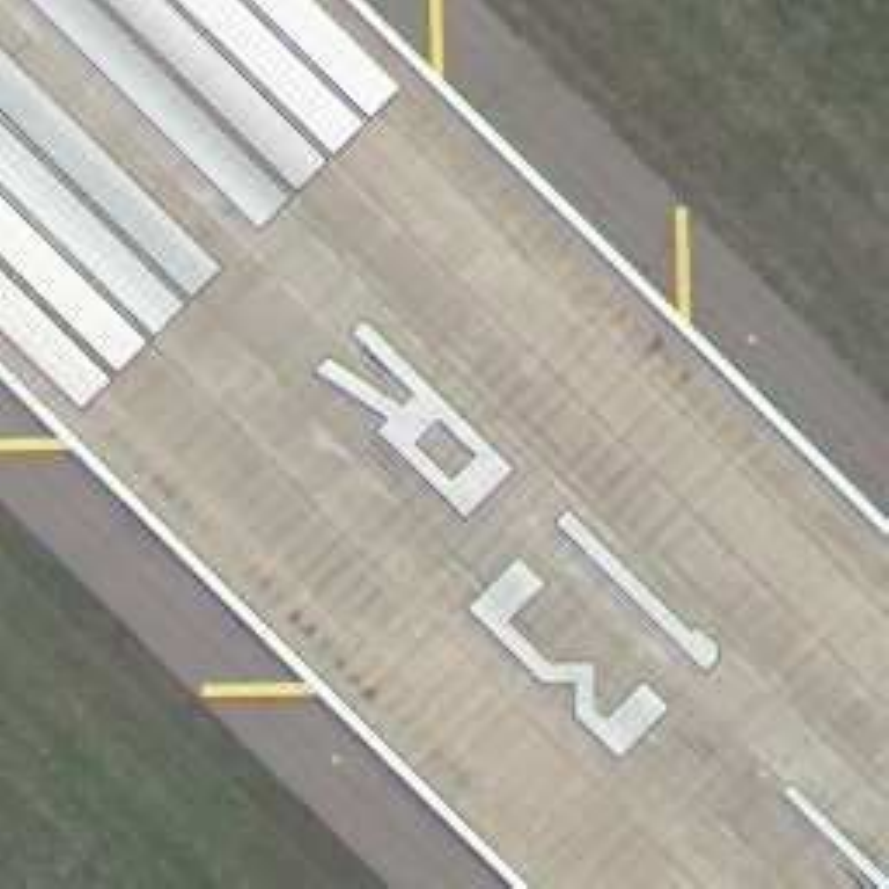}
	}
	\hspace{1mm}
	\subfloat[Airplane]{
		\includegraphics[width=\exFigSize\textwidth, keepaspectratio=true]{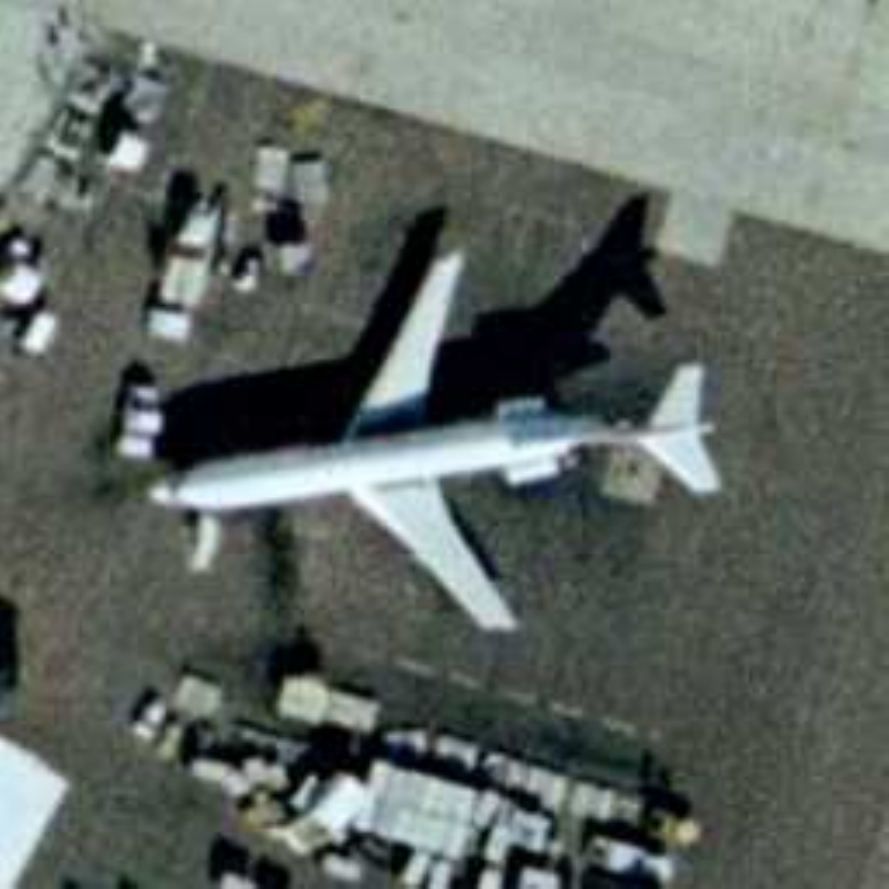}
		\includegraphics[width=\exFigSize\textwidth, keepaspectratio=true]{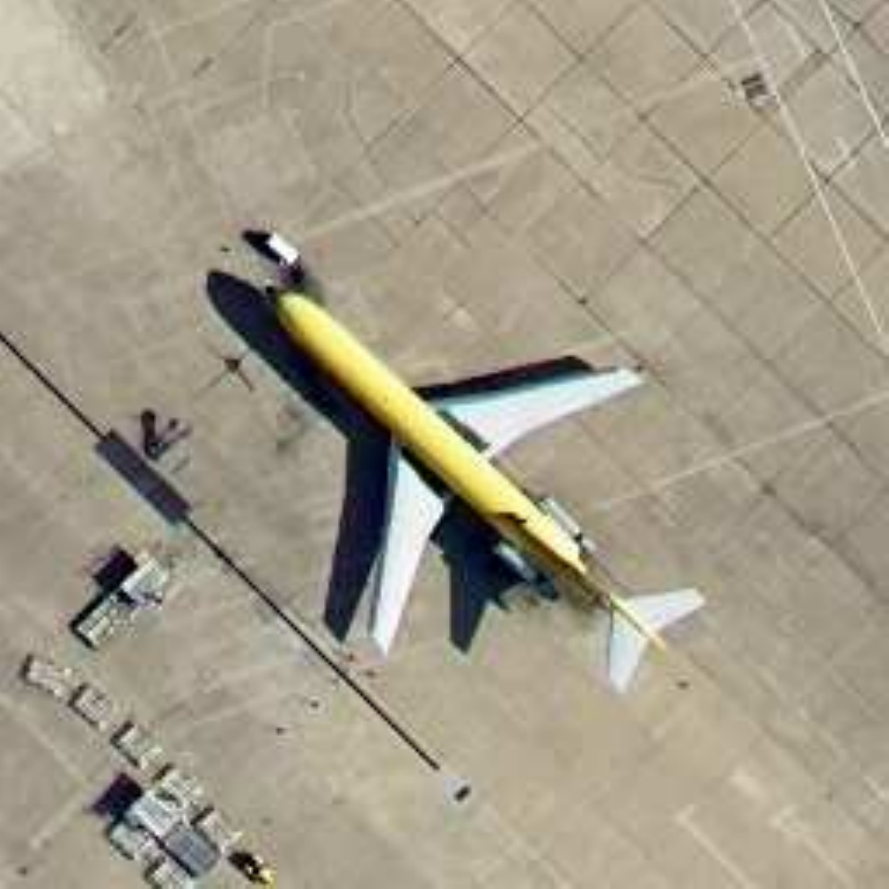}
	}
	\caption{Examples of the UCMerced Land Use Dataset.}
	\label{fig:merced_dataset}
\end{figure}

\subsubsection{RS19 Dataset}  \label{subsubsec:rs19}

This public 
dataset~\cite{xia2010structural} contains 1,005 high-spatial resolution images with $600\times600$ pixels divided into 19 classes, with approximately 50 images per class. 
Exported from Google Earth, which provides high-resolution satellite images up to half a meter, this dataset has samples collected from different regions all around the world, which increases its diversity but creates challenges due to the changes in resolution, scale, orientation and illuminations of the images.
The 19 classes, including: airport, beach, bridge, river, forest, meadow, pond, parking, port, viaduct, residential area, industrial area, commercial area, desert, farmland, football field, mountain, park and railway station.
Figure~\ref{fig:rs19_dataset} presents examples of some classes.

\begin{figure}[t!]
	\centering
	\scriptsize
	\subfloat[Industrial]{
		\includegraphics[width=\exFigSize\textwidth, keepaspectratio=true]{industrial_14.jpg}
		\includegraphics[width=\exFigSize\textwidth, keepaspectratio=true]{industrial_24.jpg}
	}
	\hspace{1mm}
	\subfloat[Commercial]{
		\includegraphics[width=\exFigSize\textwidth, keepaspectratio=true]{commercial_01.jpg}
		\includegraphics[width=\exFigSize\textwidth, keepaspectratio=true]{commercial_46.jpg}
	}
	\hspace{1mm}
	\subfloat[Park]{
		\includegraphics[width=\exFigSize\textwidth, keepaspectratio=true]{Park_22.jpg}
		\includegraphics[width=\exFigSize\textwidth, keepaspectratio=true]{Park_42.jpg}
	}
	\hspace{1mm}
	\subfloat[Residential]{
		\includegraphics[width=\exFigSize\textwidth, keepaspectratio=true]{residential_24.jpg}
		\includegraphics[width=\exFigSize\textwidth, keepaspectratio=true]{residential_34.jpg}
	}
	\hspace{1mm}
	\subfloat[Viaduct]{
		\includegraphics[width=\exFigSize\textwidth, keepaspectratio=true]{viaduct_13.jpg}
		\includegraphics[width=\exFigSize\textwidth, keepaspectratio=true]{viaduct_28.jpg}
	}
	\hspace{1mm}
	\subfloat[Bridge]{
		\includegraphics[width=\exFigSize\textwidth, keepaspectratio=true]{bridge_16.jpg}
		\includegraphics[width=\exFigSize\textwidth, keepaspectratio=true]{bridge_33.jpg}
	}
	\caption{Examples of the RS19 Dataset.}
	\label{fig:rs19_dataset}
\end{figure}

\subsubsection{Brazilian Coffee Scenes} \label{subsubsec:coffee_dataset}

\begin{figure}[t!]
	\centering
	\scriptsize
	\subfloat[Coffee]{
		\includegraphics[width=\exFigSize\textwidth]{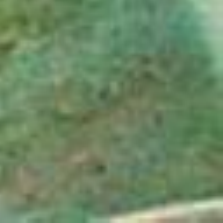}
		\includegraphics[width=\exFigSize\textwidth]{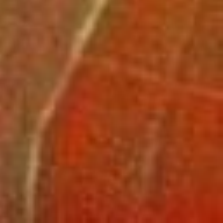}
		\includegraphics[width=\exFigSize\textwidth]{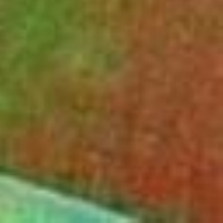}
	}
	\hspace{1mm}
	\subfloat[Non-coffee]{
		\includegraphics[width=\exFigSize\textwidth]{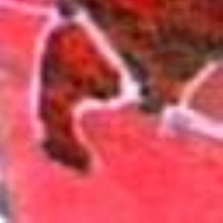}
		\includegraphics[width=\exFigSize\textwidth]{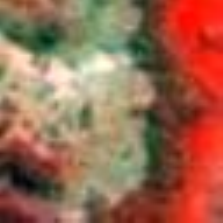}
		\includegraphics[width=\exFigSize\textwidth]{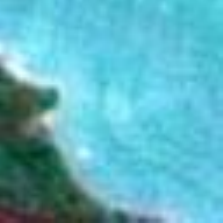}
	}
	
	\caption{Examples of coffee and non-coffee samples in the Brazilian Coffee Scenes dataset. The similarity among samples of opposite classes is notorious. The intraclass variance is also perceptive.}
	\label{fig:coffee}
\end{figure}

This dataset~\cite{penattideep} is composed of multi-spectral scenes taken by the SPOT sensor in 2005 over four counties in the State of Minas Gerais, Brazil: Arceburgo, Guaran\'esia, Guaxup\'e, and Monte Santo.
Images of each county were partitioned into multiple tiles of $64\times64$ pixels, which generated 2,876 images equally divided into 2 classes (coffee and non-coffee). Some samples of this dataset are presented in Figure~\ref{fig:coffee}. 
It is important to emphasize that these images are composed of green, red, and near-infrared bands, which are the most useful and representative ones for discriminating vegetation areas.

This dataset is very challenging for several different reasons: (i) high intraclass variance, caused by different crop management techniques, (ii) scenes with different plant ages, since coffee is an evergreen culture and, (iii) images with spectral distortions caused by shadows, since the South of Minas Gerais is a mountainous region.

\subsection{Classical Feature Extraction Strategies} \label{subsec:descriptors}

Several previously state-of-the-art descriptors have been selected to be evaluated, based on preceding works~\cite{Yang2008:ICIP,dosSantos2010:VISAPP, SandeEvaluatingTPAMI2010,PenattiComparativeJVCIR2012, Yang2013:TGRS,dosSantos2014:JSTARS}, in which they were evaluated for remote sensing image classification, texture and color image retrieval/classification, and web image retrieval. 
Our selection includes simple global low-level descriptors, like descriptors based on color histograms and variations, and also descriptors based on bags of visual words (BoVW).

\subsubsection{Low-Level descriptors} \label{subsubsec:lowlevel}

There is a myriad of descriptors available in the literature~\cite{PenattiComparativeJVCIR2012} that can be used to represent visual elements.
Clearly, different descriptors may provide distinct information about images producing contrastive results.
Thus, we tested a diverse set of 7 descriptors based on color, texture, and gradient properties, in order to extract visual features from each image and evaluate the potential of deep features when compared them.
The global low-level descriptors considered are: 
Auto-Correlogram Color (\textbf{ACC})~\cite{HuangColorCorrelogram1997}, 
Border/Interior Pixel Classification (\textbf{BIC})~\cite{StehlingBIC2002}, 
Local Color Histogram (\textbf{LCH})~\cite{SwainColor}, 
Local Activity Spectrum (\textbf{LAS})~\cite{TaoLAS2000}, 
Statistical Analysis of Structural Information (\textbf{SASI})~\cite{VuralSASI2003}, 
Histogram of Oriented Gradients (\textbf{HOG})~\cite{DalalHOGCVPR2005}, and
\textbf{GIST}~\cite{OlivaGIST2001}.


The implementations of ACC, BIC, LCH, SASI, and LAS descriptors follow the specifications of~\cite{PenattiComparativeJVCIR2012}.
GIST implementation is the one used in~\cite{DouzeGistCIVR2009} with the parameters discussed therein.\footnote{\url{http://lear.inrialpes.fr/software} (as of March 14th, 2015).}
The implementation of Histogram of Oriented Gradients (HOG) was obtained from the VLFeat framework~\cite{Vedaldi08VLfeat}. 
We used HOG in different configurations, varying the cell size in $14\times14$, $20\times20$, $40\times40$ and $80\times80$ pixels, but keeping the orientation binning in 9 bins.

\subsubsection{Mid-Level descriptors} \label{subsubsec:midlevel}

Bag of visual words (\textbf{BoVW}) and their variations~\cite{SivicVideoGoogle2003,GemertVisuaWordAmbiguityPAMI2010, BoureauMidLevelCVPR2010,LazebnikBeyondBags2006,PenattiWSAPR2013,AvilaBossaNovaCVIU2013,PerronninFisherECCV2010} are considered mid-level representations, since these methods create a codebook of visual discriminating patches (visual words), and then, compute statistics (using the codebook) about the visual word occurrences in the test image.
BoVW descriptors have been the state of the art for several years in the computer vision community and are still important candidates to perform well in many tasks.

We tested BoVW in several different configurations:
\begin{itemize}
	\item Sampling: sparse (Harris-Laplace detector) or dense (grid of circles with 6 pixels of radius)
	\item Low-level descriptor: SIFT and OppponentSIFT~\cite{SandeEvaluatingTPAMI2010}
	\item Visual codebooks of size: 100, 1000, 5000, and 10000
	\item Assignment: hard or soft (with $\sigma$ = 90 or 150)
	\item Pooling: average, max pooling or WSA~\cite{PenattiWSAPR2013}
\end{itemize}

To differentiate them in the experiments, we used the following naming:
$BX^{w}_{cp}$, where
$X$ is S (sparse sampling) or D (dense sampling);
$w$ is the codebook size;
$c$ refers to the coding scheme used, h (hard), s (soft);
$p$ refers to the pooling technique used, a (average), m (max), or W (WSA).

The low-level feature extraction of BoVW descriptors was based on the implementation of van de Sande et al.~\cite{vandeSandeITM2011}.
For BoVW, in UCMerced and RS19 datasets, we used SIFT~\cite{LoweSIFT2004} to describe each patch, but in the Brazilian Coffee dataset, we used OpponentSIFT~\cite{SandeEvaluatingTPAMI2010}, as color should provide more discriminating power. 



\subsection{ConvNets} \label{subsec:cnn}

As the main goal of the paper is to evaluate the strategies for better exploiting existing deep ConvNets, we selected some of the most popular ConvNets available nowadays.
All networks presented in this section, except for the OverFeat ones, were implemented in Convolutional Architecture for Fast Feature Embedding~\cite{jia2014caffe}, or simply Caffe, a fully open-source framework that affords clear and easy implementations of deep architectures.
The OverFeat ConvNets were proposed and implemented by the namesake framework~\cite{sermanet-iclr-14}.

\begin{figure}[t!]
	\centering
	\scriptsize
	\subfloat[PatreoNet]{
		\includegraphics[width=\textwidth]{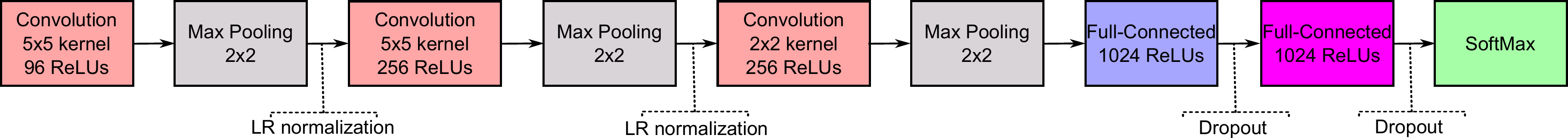}
		\label{fig:patreonet}
	}
	\hspace{3mm}
	\subfloat[AlexNet]{
		\includegraphics[width=\textwidth]{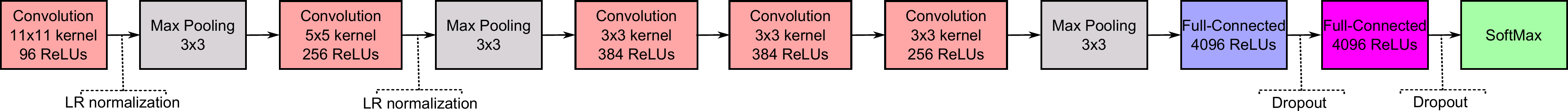}
		\label{fig:alexnet}
	}
	\hspace{3mm}
	\subfloat[CaffeNet]{
		\includegraphics[width=\textwidth]{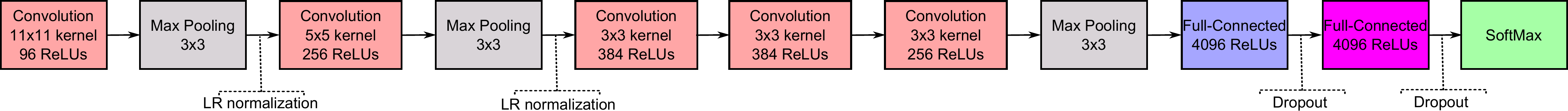}
		\label{fig:caffenet}
	}
	\hspace{3mm}
	\subfloat[$VGG_{16}$]{
		\includegraphics[width=\textwidth]{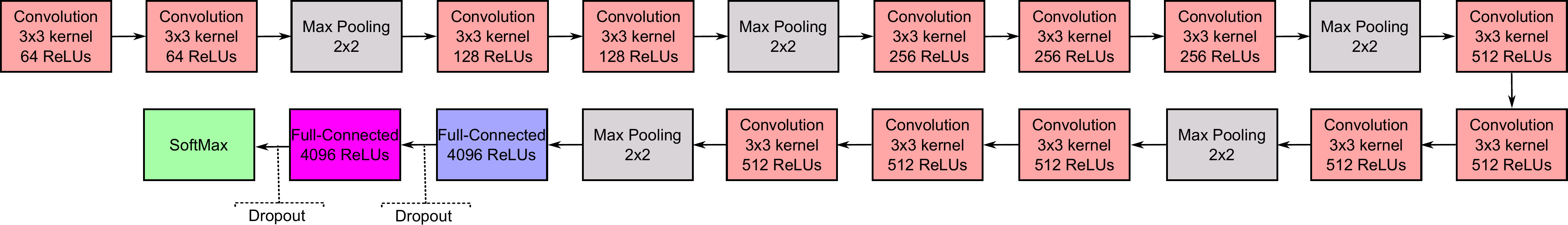}
		\label{fig:vgg16}
	}
	\hspace{3mm}
	\subfloat[OverFeat$_{S}$]{
		\includegraphics[width=\textwidth]{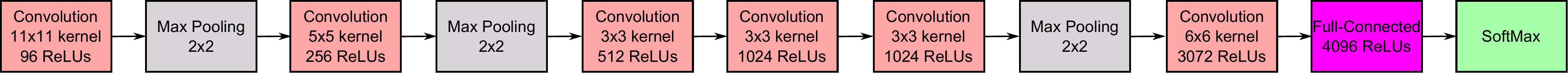}
		\label{fig:overfeat_s}
	}
	\hspace{3mm}
	\subfloat[OverFeat$_{L}$]{
		\includegraphics[width=\textwidth]{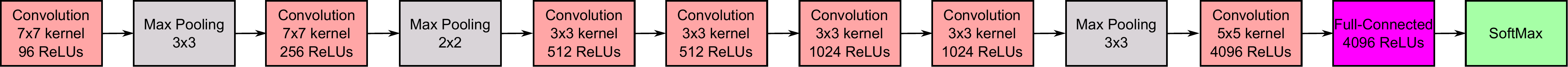}
		\label{fig:overfeat_l}
	}
	\hspace{3mm}
	\subfloat[GoogLeNet]{
		\includegraphics[width=\textwidth]{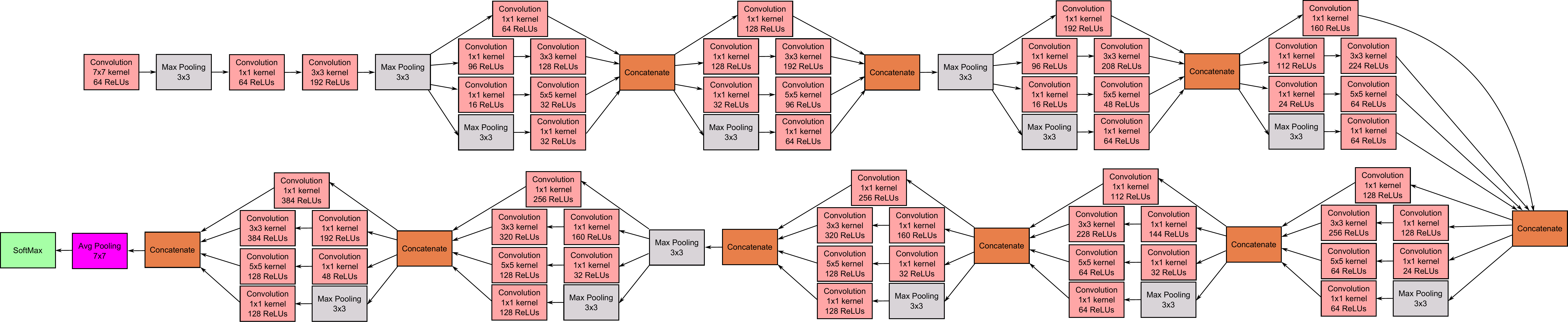}
		\label{fig:googlenet}
	}
	\caption{Architectures of different the ConvNets evaluated in this work.
		Purple boxes indicate the layers from where features were extracted in the case of using the ConvNets as feature extractors.}
	\label{fig:convnets}
\end{figure}

Figure~\ref{fig:convnets} presents an overview of the selected ConvNet architectures.
Some information about the networks evaluated in this work are presented in Table~\ref{tab:compare_convnet}.
GoogLeNet~\cite{szegedy2014going} is the biggest network, with higher number of connections, followed OverFeat$_L$ and $VGG_{16}$.
In fact, these networks, OverFeat and $VGG_{16}$, are also the ones with higher number of parameters, which require more memory during the training process.
We describe some properties of each ConvNet in the following subsections.

\begin{table}[]
	\centering
	\caption{Some statistics about the deep networks evaluated in this work.}
	\label{tab:compare_convnet}
	\begin{tabular}{@{}lrr@{}}
		\toprule
		\multicolumn{1}{c}{\textbf{\begin{tabular}[c]{@{}c@{}}Networks \end{tabular}}} &
		\multicolumn{1}{c}{\textbf{\begin{tabular}[c]{@{}c@{}}\#parameters\\ (millions)\end{tabular}}} & \multicolumn{1}{c}{\textbf{\begin{tabular}[c]{@{}c@{}}\#connections\\ (millions)\end{tabular}}} \\
		\midrule
		\textbf{OverFeat$_S$} & 145 & 2,810 \\
		\textbf{OverFeat$_L$} & 144 & 5,369 \\
		\textbf{AlexNet}      & 60  & 630   \\
		\textbf{CaffeNet}     & 60  & 630   \\
		\textbf{GoogLeNet}    & 5   & \textgreater10,000 \\
		\textbf{VGG$_{16}$}   & 138 & 4,096 \\
		\textbf{PatreoNet}    & 15  & 200 \\
		\bottomrule
	\end{tabular}
\end{table}

\subsubsection{PatreoNet} \label{subsubsec:patreonet}

PatreoNet, presented in~\cite{nogueira2015improving}, is a network capable of learning specific spatial features from remote sensing images, without any pre-processing step or descriptor evaluation.
This network, which architecture can be seen in Figure~\ref{fig:patreonet}, has 3 convolutional layers, 3 pooling ones and 3 fully-connected ones (considering the softmax).
PatreoNet was only used in full-training strategy, since it has no model pre-trained in large datasets.

\subsubsection{AlexNet} \label{subsubsec:alexNet}

\textbf{AlexNet}, proposed by Krizhevsky et al.~\cite{krizhevsky2012imagenet}, was the winner of ImageNet Large Scale Visual Recognition Challenge (ILSVRC)~\cite{deng2009imagenet} in 2012.
This ConvNet, that has 60 million parameters and 650,000 neurons, consists of five convolutional layers, some of which are followed by max-pooling layers, and three fully-connected layers with a final softmax.
Its final architecture can be seen in Figure~\ref{fig:alexnet}.
It was a breakthrough work since it was the first to employ non-saturating neurons, GPU implementation of the convolution operation and dropout to prevent overfitting.

In the experiments, AlexNet was used as a feature extractor network by extracting the features from the last fully-connected layer (purple one in Figure~\ref{fig:alexnet}), which results in a feature vector of 4,096 dimensions.
AlexNet was also fine-tuned for all the datasets used in the experiments.
For UCMerced and RS19 datasets the network was fine-tuned by giving more importance to the final softmax layer (without freezing any layer), while for the Coffee Dataset, the first three layers were frozen and the final ones participate normally in the learning process.
Finally, AlexNet was fully trained from scratch for all datasets under the same configurations of the original model~\cite{krizhevsky2012imagenet}.

\subsubsection{CaffeNet} \label{subsubsec:caffeNet}

\textbf{CaffeNet}~\cite{jia2014caffe} is almost a replication of ALexNet~\cite{krizhevsky2012imagenet} with some important differences:
(i) training has no relighting data-augmentation, and
(ii) the order of pooling and normalization layers is switched (in CaffeNet, pooling is done before normalization).
Thus, this network, which architecture can be seen in Figure~\ref{fig:caffenet}, has the same number of parameters, neurons and layers of the AlexNet.
Given its similarity to the AlexNet architecture, in the experiments, CaffeNet were exploited in the same way of the aforementioned network.

\subsubsection{GoogLeNet} \label{subsubsec:googleNet}

\textbf{GoogLeNet}, presented in~\cite{szegedy2014going}, is the ConvNet architecture that won the ILSVRC-2014 competition (classification and detection tracks).
Its main peculiarity is the use of inception modules, which reduce the complexity of the expensive filters of traditional architectures allowing multiple filter, with different resolutions, to be used in parallel.
GoogLeNet has two main advantages: (i) utilization of filters of different sizes at the same layer, which maintains more spatial information, and (ii) reduction of the number of parameters of the network, making it less prone to overfitting and allowing it to be deeper.
In fact, GoogLeNet has 12 times fewer parameters than AlexNet, i.e., approximately 5 millions of parameters.
Specifically, the 22-layer GoogLeNet architecture, which can be seen in Figure~\ref{fig:googlenet}, has more than 50 convolutional layers distributed inside the inception modules.

In our experiments, 
GoogLeNet was used as a feature extractor network by extracting the features from the last pooling layer (purple one in Figure~\ref{fig:googlenet}), which results in a feature vector of 1,024 dimensions.
GoogLeNet was fine-tuned for all datasets just like AlexNet, with exact the same strategies for each dataset.
Finally, GoogLeNet was fully trained from scratch for all datasets under the same configurations of the original model~\cite{szegedy2014going}.

\subsubsection{VGG ConvNets} \label{subsubsec:vgg}

\textbf{VGG} ConvNets, presented in~\cite{simonyan2014very}, won the localization and classification tracks of the ILSVRC-2014 competition.
Several networks have been proposed in this work, but two have become more successful: $VGG_{16}$ and $VGG_{19}$.
Giving the similarity of both networks, we choose to work with the former one because of its simpler architecture and slightly better results.
However, similar results obtained by $VGG_{16}$ should be also yielded by $VGG_{19}$.
This network, which architecture can be seen in Figure~\ref{fig:vgg16}, has 13 convolutional layers, 5 pooling ones and 3 fully-connected ones (considering the softmax).

$VGG_{16}$ was used as a feature extractor network by extracting the features from the last fully-connected layer (purple one in Figure~\ref{fig:vgg16}), which results in a feature vector of 4,096 dimensions.
$VGG_{16}$ was also fine-tuned in the UCMerced and RS19 datasets by giving more importance to the final softmax layer, without freezing any layer.
However, this network could not be fine-tuned for the Brazilian Coffee dataset (the most different one) as well as could not be fully trained from scratch for any dataset.
This problem is due to the large amount of memory required by this network, as presented in Table~\ref{tab:compare_convnet}, which allows only small values of batch size to be used during the training process.
Since larger values of batch size, combined with other parameters (weight decay, learning rate), help the convergence of a ConvNet in the training process~\cite{deeplearningbook,bengio2009learning}, there was no convergence in the aforementioned scenarios.

\subsubsection{OverFeat ConvNets} \label{subsubsec:overfeat}


\textbf{OverFeat}~\cite{OverFeatIntegrated2014}, a deep learning framework focused on ConvNets and winner of the detection track of ILSVRC 2013, has two ConvNet models available, both trained on the ImageNet 2012 training set~\cite{ILSVRCarxiv14}, which can be used to extract features and/or to classify images. 
There is a small (\textit{fast} -- OverFeat$_{S}$) and a larger network (\textit{accurate} -- OverFeat$_{L}$), both having similarities with AlexNet~\cite{krizhevsky2012imagenet}.
The main differences are: (i) no response normalization, and (ii) non-overlapping pooling regions. 
OverFeat$_{L}$, which architecture can be seen in Figure~\ref{fig:overfeat_l}, has more differences including:
(i) one more convolutional layer and,
(ii) the number and size of feature maps, since different number of kernels and stride were used for the convolutional and the pooling layers. 
In the other way around, OverFeat$_{S}$, which architecture can be seen in Figure~\ref{fig:overfeat_s}, is more similar to the AlexNet, differing only in the number and size of feature maps.
The main differences between the two OverFeat networks are the stride of the first convolution, the number of stages and the number of feature maps~\cite{OverFeatIntegrated2014}.

These ConvNets are only used as feature extractor, since no model was provided in order to perform fine-tuning or full-training of the networks.
Considering this strategy, a feature vector of 4,096 dimensions is obtained from the last fully-connected layer, which 
are illustrated as 
purple layers in Figures~\ref{fig:overfeat_s} and~\ref{fig:overfeat_l}, for OverFeat$_{S}$ and OverFeat$_{L}$, respectively.

\subsection{Experimental protocol} \label{subsec:protocol}

We carried out all experiments with a 5-fold cross-validation protocol.
Therefore, the dataset was arranged into five folds with almost same size, i.e., the images are almost equally divided into five non-overlapping sets.
Specifically, the UCMerced and RS19 datasets have 5 folds, unbalanced in terms of the number of samples per class, with 420 and 201 images, respectively.
For the Brazilian Coffee Scenes dataset, 4 folds have 600 images each and the 5$^{th}$ has 476 images, all folds are balanced with coffee and non-coffee samples (50\% each).

When performing fine-tuning or training a network for scratch, at each run, three folds are used as training-set, one as validation (used to tune the parameters of the network) and the remaining one is used as test-set.
It is important to mention that when changing the folds which are train, validation and test (during the cross-validation process), the full training or fine tuning of the network starts from the beginning. Five different networks are obtained, one for each step of the 5-fold cross-validation process.
That is, there is no contamination of the training set with testing data. 
When using the ConvNets as feature extractors, four sets are used as training while the last is the test-set.
Still considering this strategy, we always used linear SVM as the final classifier.

When fine-tuning or full-training a network, we basically preserve the parameters of the original author, varying only two according to Table~\ref{tab:parameters}.
It is important to highlight that there is no training when using a pre-trained ConvNet (without fine-tuning) as feature extractor, thus there no parameters to vary.
\begin{table}[]
	\centering
	\caption{Parameters utilized in fine-tuning and full-training strategies.}
	\label{tab:parameters}
	\begin{tabular}{@{}lrr@{}}
		\toprule
		\textbf{Strategy}      & \multicolumn{1}{l}{\textbf{\#iterations}} & \multicolumn{1}{l}{\textbf{Learning Rate}} \\
		\midrule
		\textbf{Fine-tuning}   & 20,000                                    & 0.001                                      \\
		\textbf{Full-training} & 50,000                                    & 0.01                                       \\
		\bottomrule
	\end{tabular}
\end{table}

The results are reported in terms of average accuracy and standard deviation among the 5 folds.
For a given fold, we compute the accuracy for each class and then compute the average accuracy among the classes. 
This accuracy is used to compute the final average accuracy among the 5 folds.

All 
experiments were performed on a 64 bits Intel i7 4960X machine with 3.6GHz of clock and 64GB of RAM memory. 
Two GPUs were used: a GeForce GTX770 with 4GB of internal memory and a GeForce GTX Titan X with 12GB of memory, both under a 7.5 CUDA version. 
Ubuntu version 14.04.3 LTS was used as operating system.

\section{Results and Discussion}  \label{sec:results}

In this section, we present and discuss the experimental results. Firstly, we discuss the power of generalization of ConvNets as feature descriptors and compare them with low-level and mid-level representations (Section~\ref{subsec:generalization}). Then, we compare the performance of the three different strategies for exploiting the existing ConvNets (Section~\ref{subsec:exploitConvNets}). Finally, we compare the most accurate ConvNets against some of the state-of-the-art methods for each dataset (Section~\ref{subsec:stateComparison}).


\subsection{Generalization Power Evaluation} \label{subsec:generalization}

In this subsection, we compare six pre-trained ConvNets used as descriptors against low-level and mid-level representations for aerial and remote sensing scene classification. We conducted several experiments in order to evaluate the best BoVW configurations, but only the top-5 best were reported. It is also important to highlight that the ConvNets results in this subsection refer to their use as feature extractors, by considering the features of the last layer before softmax as input for a linear SVM. So, the original network was not used to classify the samples. The objective is to observe how deep features perform in datasets from different domains they were trained.

In Figure~\ref{fig:accuracy_ucmerced}, we show the average accuracy of each descriptor and ConvNet for the UCMerced dataset.
We can notice that ConvNet features achieve the highest accuracy rates ($\geq$ 90\%).
CaffeNet, AlexNet, and $VGG_{16}$ yield the highest average accuracies (more than 93\%).  
GoogLeNet achieves 92.80\% $\pm$ 0.61 and OverFeat achieves 90.91\% $\pm$ 1.19 for the small and 90.13\% $\pm$ 1.81 for the large network.
SASI is the best global descriptor (73.88\% $\pm$ 2.25), while the best BoVW configurations are based on dense sampling, 5 or 10 thousand visual words and soft assignment with max pooling ($\sim$81\%).

\newcommand{\resultsFigSize}{0.7}
\begin{figure}[ht!]
	\centering
	\includegraphics[trim = 0mm 2mm 0mm 28mm, clip, width=\resultsFigSize\textwidth]{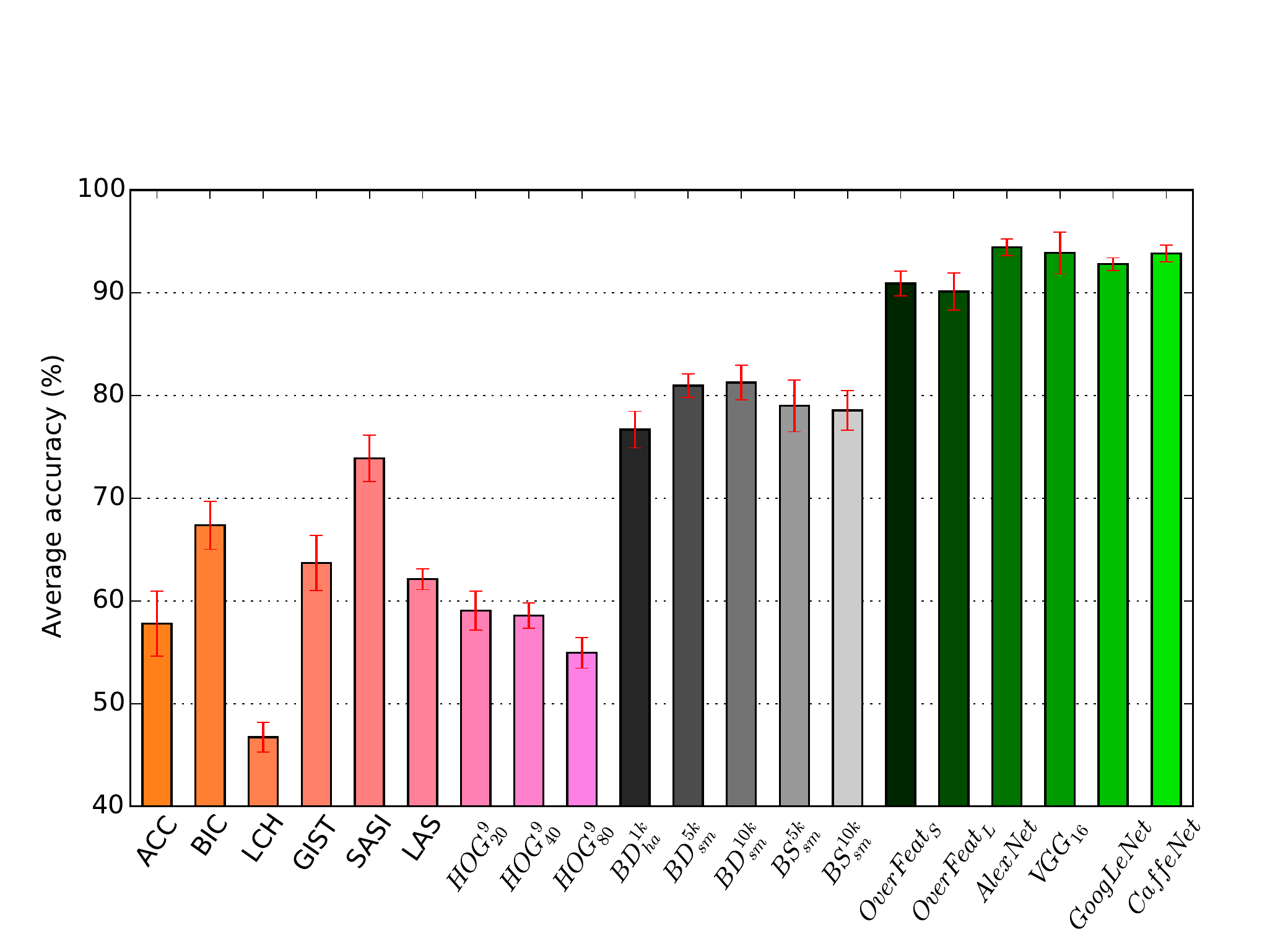}
	\caption{Average accuracy of pre-trained ConvNets used as feature extractors and low- and mid-level descriptors for the UCMerced Land-use Dataset.}
	\label{fig:accuracy_ucmerced}
\end{figure}

\begin{figure}[ht!]
	\centering
	\includegraphics[trim = 0mm 2mm 0mm 28mm, clip, width=\resultsFigSize\textwidth]{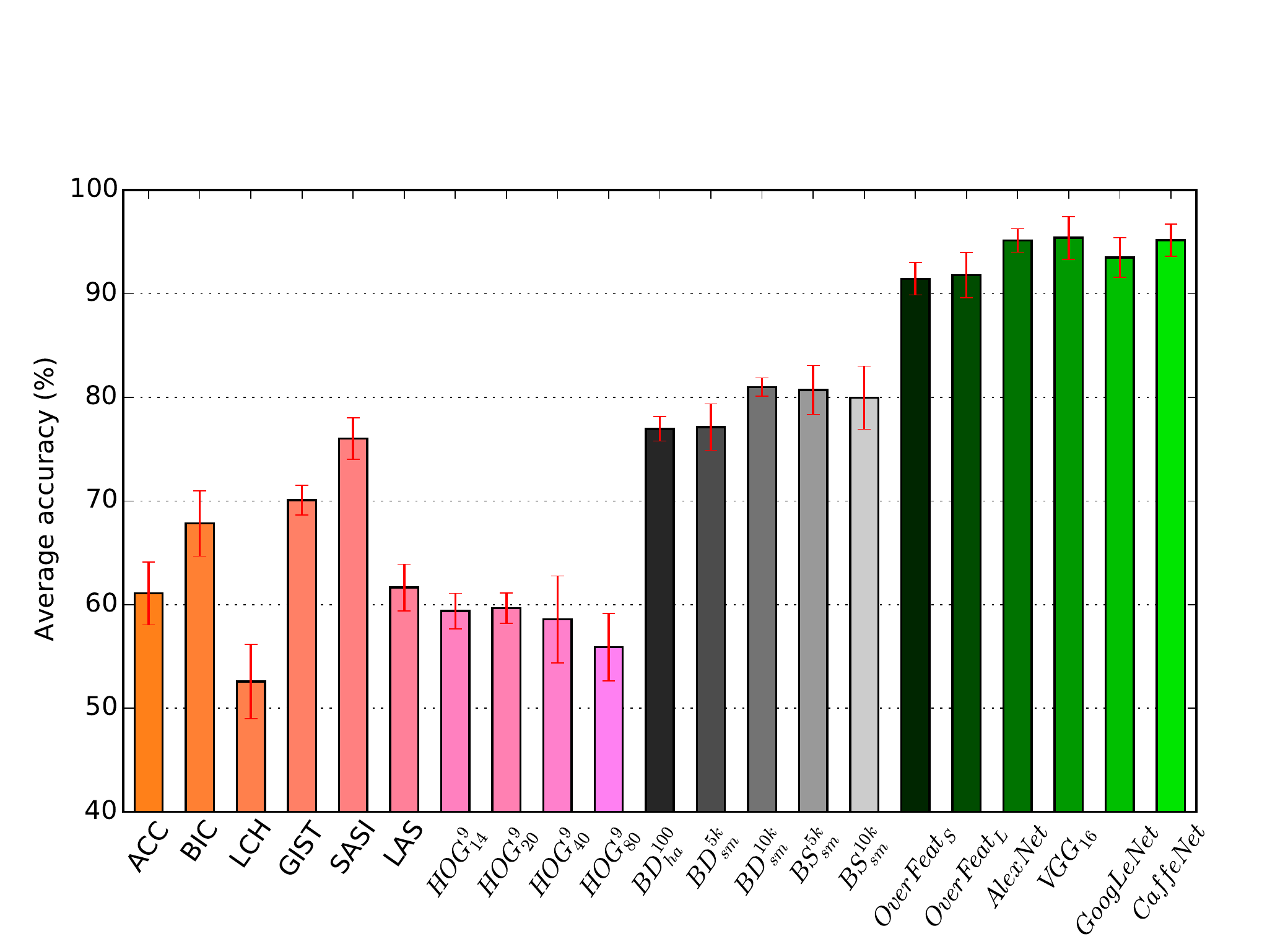}
	\caption{Average accuracy of pre-trained ConvNets used as feature extractors and low- and mid-level descriptors for the RS19 Dataset.}
	\label{fig:accuracy_rs19}
\end{figure}

In Figure~\ref{fig:accuracy_rs19}, we show the average accuracies for the RS19 dataset.
ConvNets again achieved the best results ($\geq$ 90\%). 
The best global descriptor was again SASI and the best BoVW configurations have 5 or 10 thousand visual words with soft assignment and max pooling, but in this dataset, sparse sampling also achieved similar accuracies to dense sampling.

The results with UCMerced and RS19 datasets illustrate the capacity of ConvNet features to generalize to the aerial domain.

\begin{figure}[ht!]
	\centering
	\includegraphics[trim = 0mm 2mm 0mm 28mm, clip, width=\resultsFigSize\textwidth]{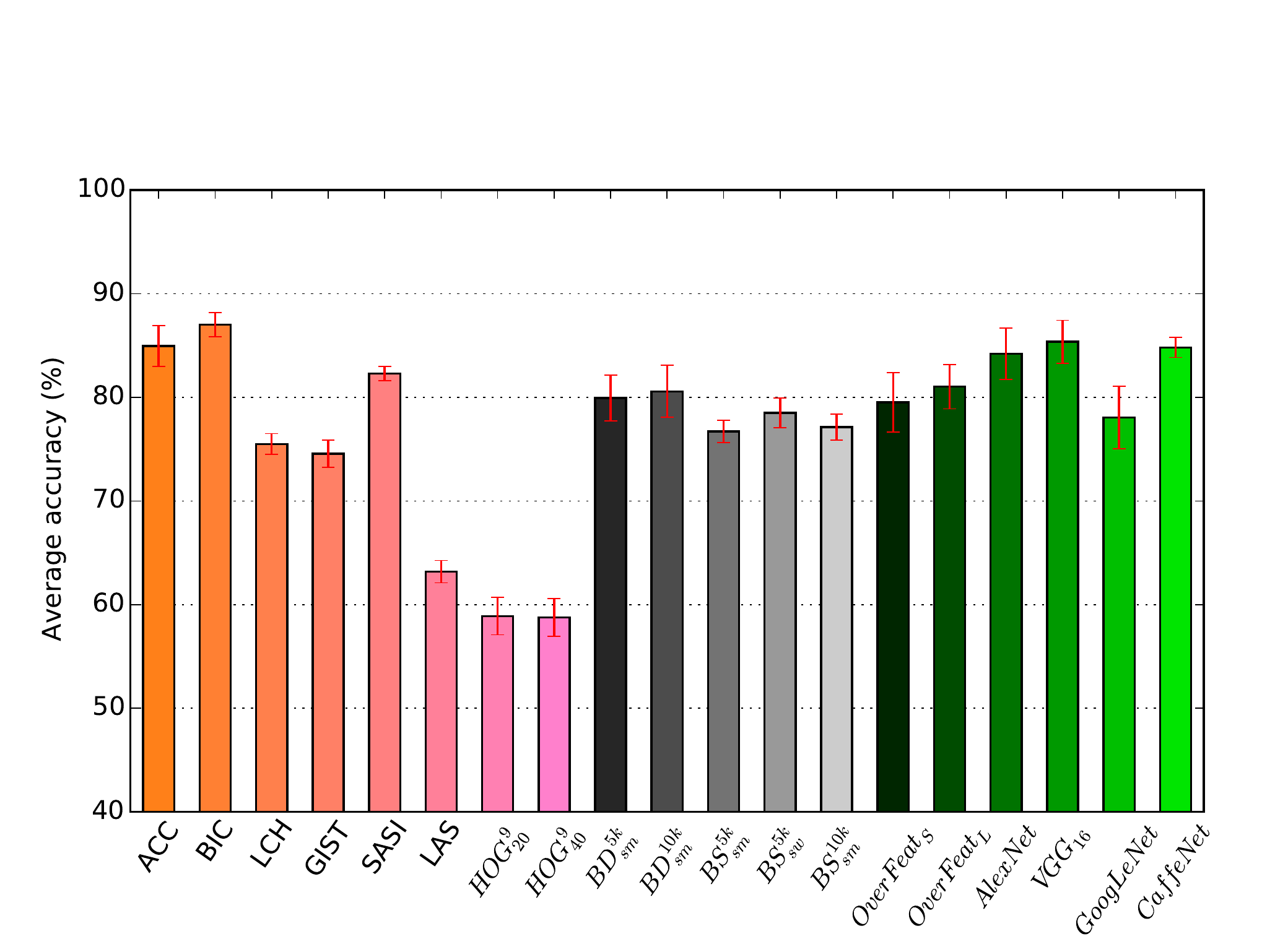}
	\caption{Average accuracy of pre-trained ConvNets used as feature extractor and low- and mid-level descriptors for the Brazilian Coffee Scenes Dataset.}
	\label{fig:accuracy_coffee}
\end{figure}

In Figure~\ref{fig:accuracy_coffee}, we show the average accuracies for the Brazilian Coffee Scenes dataset. 
In this dataset, the results are different than in the other two datasets already mentioned. 
We can see that, although most of the ConvNets achieve accuracy rates above 80\%, with $VGG_{16}$ achieving 85.36\% $\pm$ 2.08\%, BIC and ACC also achieved high accuracies. 
BIC achieved the highest average accuracy (87.03\% $\pm$ 1.17) outperforming all the descriptors including the ConvNets in this dataset.
The BIC algorithm for classifying pixels in border or interior basically separates the images into homogeneous and textured regions. 
Then, a color histogram is computer for each type of pixel. 
As for the Brazilian Coffee Scenes dataset, the differences between classes may be not only in texture but also in color properties, BIC could encode well such differences. 
The best BoVW configurations are again based on dense sampling, 5 or 10 thousand visual words and soft assignment with max pooling, and they have comparable results to OverFeat.

A possible reason for the deep features to perform better in aerial dataset than in the agricultural one is due to the particular intrinsic properties of each dataset. 
The aerial datasets have more complex scenes, composed of a lot of small objects (e.g., buildings, cars, airplanes). 
Many of these objects are composed of similar visual patterns in comparison with the ones found in the dataset used to train the ConvNets, with salient edges and borders. 

Concerning the Brazilian Coffee Scenes dataset, it is composed of finer and more homogeneous textures where the patterns are much more overlapping visually and more different than everyday objects. 
The color/spectral properties are also important in this dataset, which fit with results reported in other works~\cite{dosSantos2014:JSTARS,Faria2014:JSTARS}.

\subsection{Comparison of ConvNets Strategies} \label{subsec:exploitConvNets}     

\begin{figure}[ht!]
	\centering
	\includegraphics[trim = 0mm 2mm 0mm 28mm, clip, width=\resultsFigSize\textwidth]{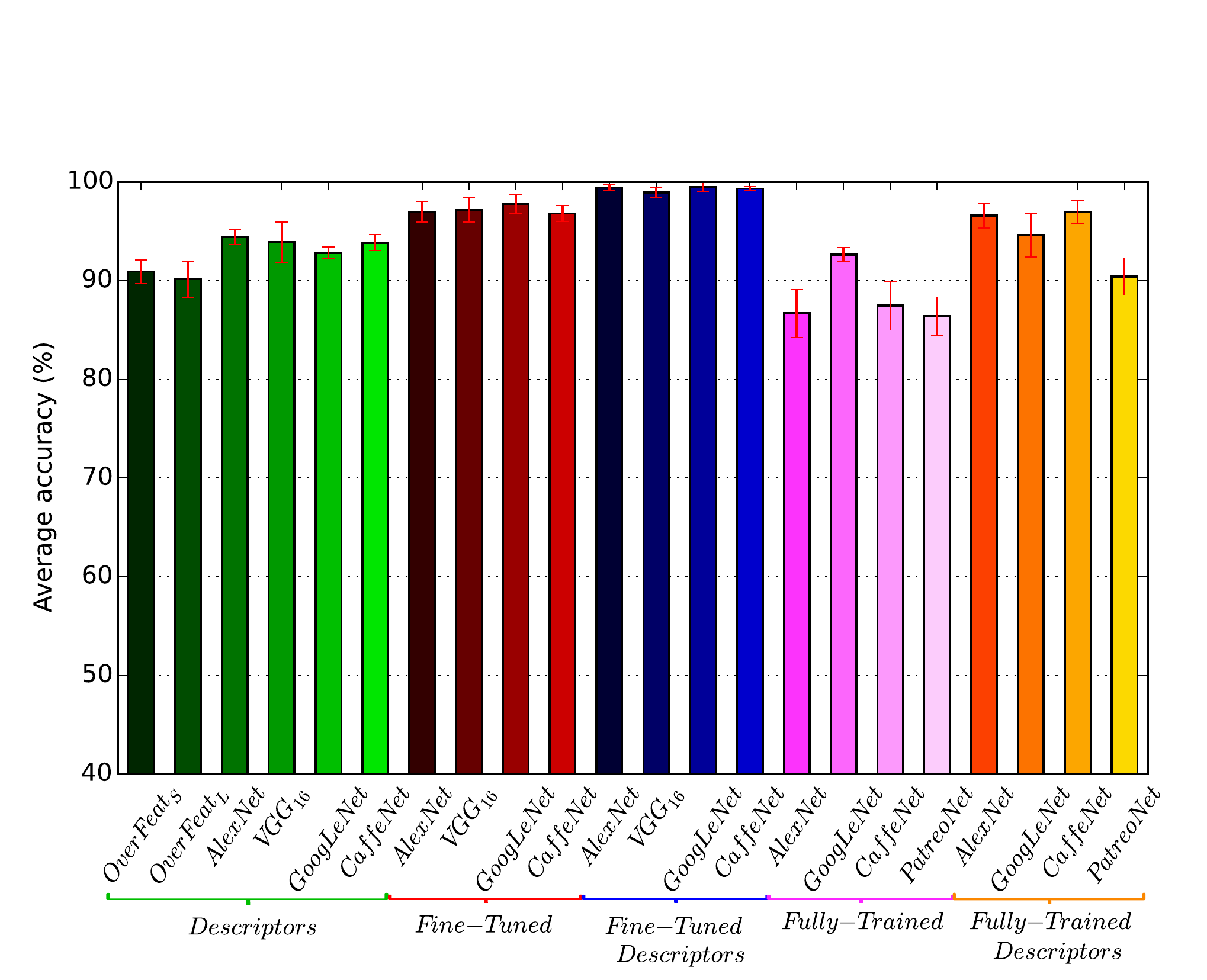}
	\caption{Average accuracy considering all possible strategies to exploit ConvNets for the UCMerced Land-use Dataset.}
	\label{fig:ucmerced_convnets}
\end{figure}

\begin{figure}[ht!]
	\centering
	\includegraphics[trim = 0mm 2mm 0mm 28mm, clip, width=\resultsFigSize\textwidth]{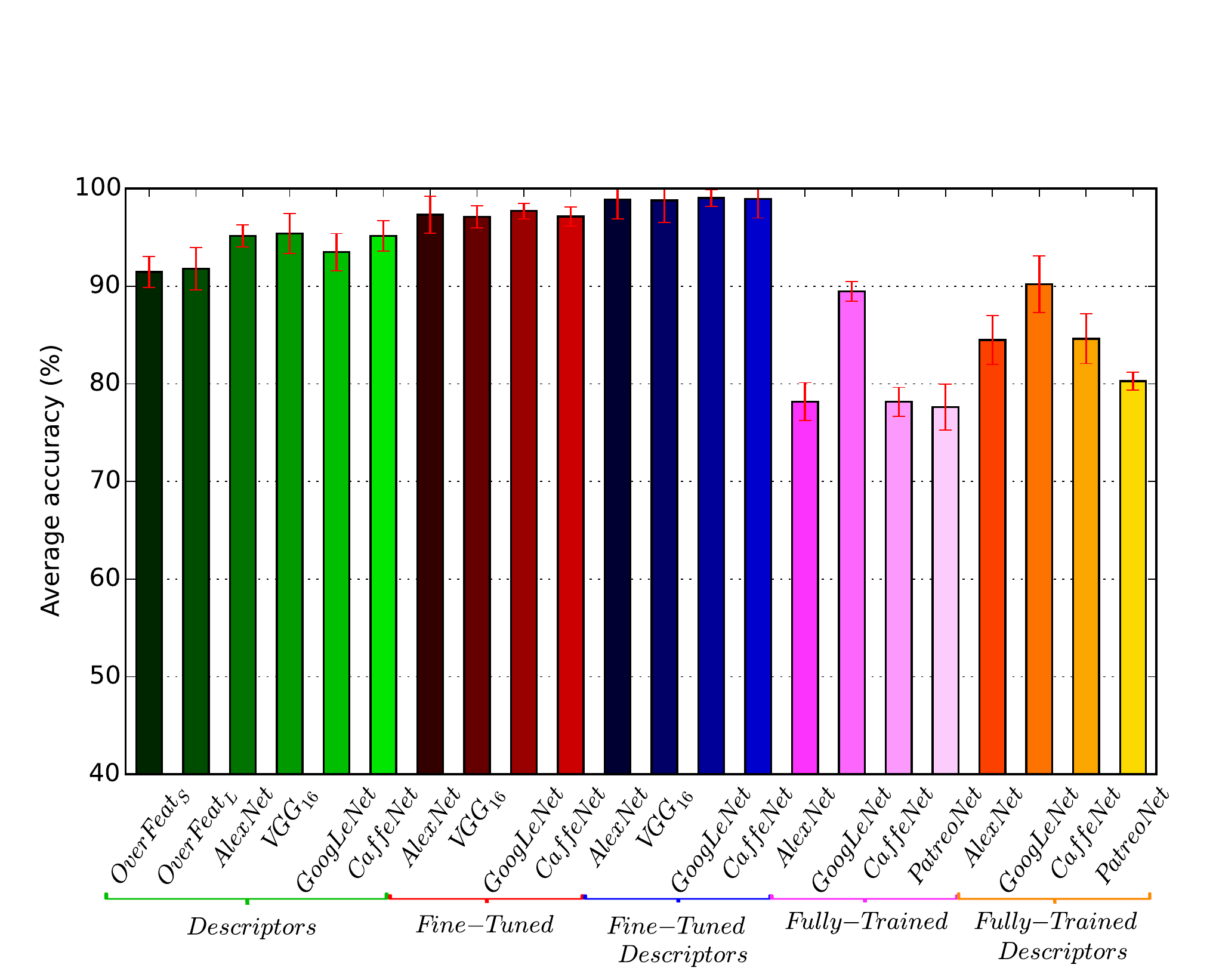}
	\caption{Average accuracy considering all possible strategies to exploit ConvNets for the RS19 Dataset.}
	\label{fig:rs19_convnets}
\end{figure}

\begin{figure}[ht!]
	\centering
	\includegraphics[trim = 0mm 2mm 0mm 28mm, clip, width=\resultsFigSize\textwidth]{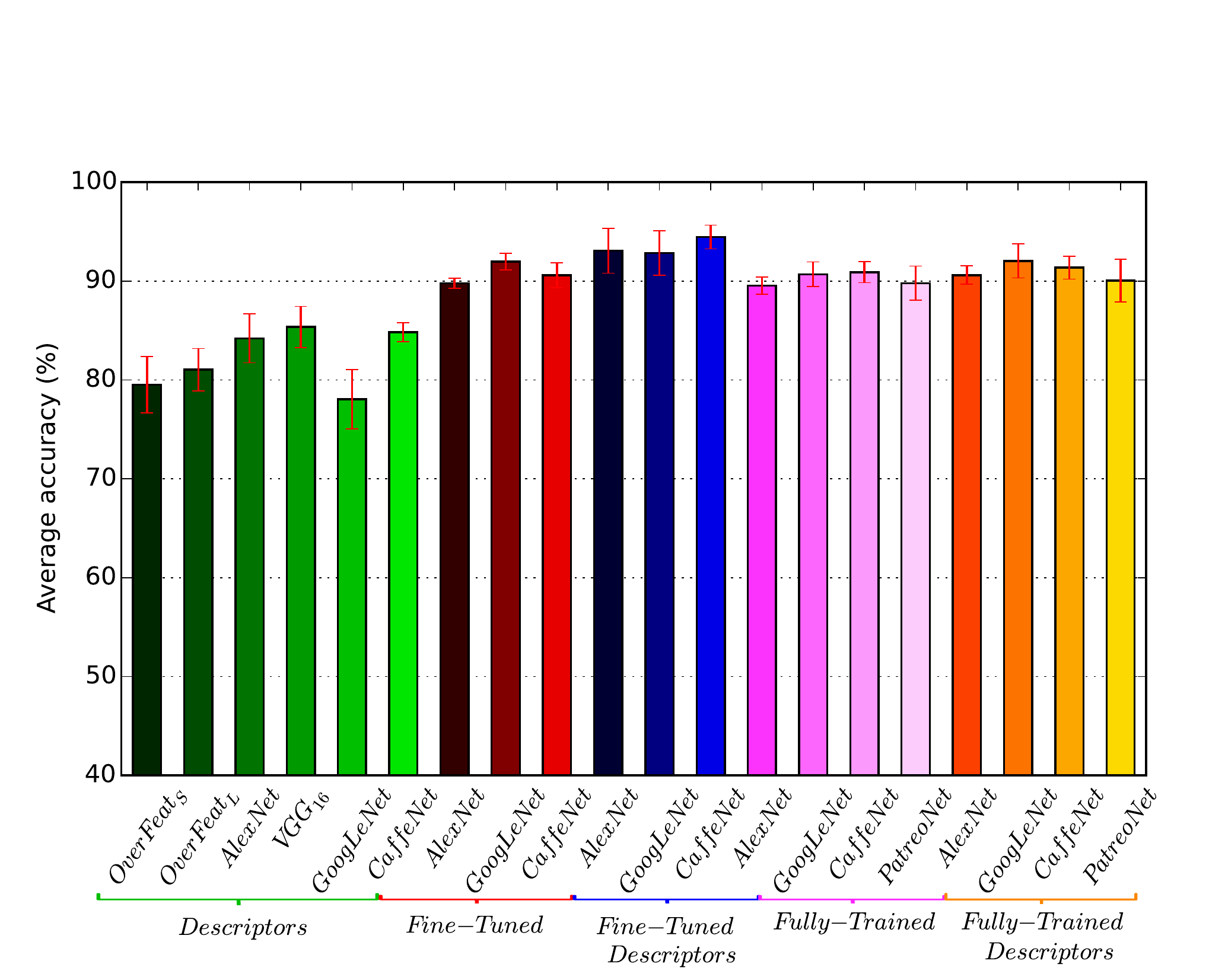}
	\caption{Average accuracy considering all possible strategies to exploit ConvNets for the Brazilian Coffee Scenes Dataset.}
	\label{fig:coffee_convnets}
\end{figure}

In this section, we compare the performance of the three different strategies for exploiting the existing ConvNets: full training, fine tuning, and using as feature extractors.
Figures~\ref{fig:ucmerced_convnets} to~\ref{fig:coffee_convnets} show the comparison of the strategies in terms of average classification accuracy.
In such figures, 
the suffix ``Descriptors'' refers to when we use the features from the last layer before the softmax layer as input for another classifier, which was a linear SVM in our case. However, it is worth mentioning that SVM was used only after the fine-tuning or the full-training process, not during the training process.

There are several interesting aspects illustrated in the graphs.
The first one is that fine tuning (red and blue bars) is usually the best strategy, outperforming the other two in all the datasets.
The difference was higher for UCMerced and RS19 datasets (Figures~\ref{fig:ucmerced_convnets} and~\ref{fig:rs19_convnets}). 
For the Coffee Scenes dataset, this difference was small, however, the Fine-Tuned Descriptors (blue bars) were slightly superior.

This advantage of fine-tuned networks, when compared to the full-trained ones, is maybe due to a better initialization in the search space.
This can be noticed in Figure~\ref{fig:convergence}, where even with more iterations, full-trained networks sticks in worse local minimum than the fine-tuned ones, what demonstrates that a better initialization of the filter weights tends to provide better results.

\newcommand{\convergenceFigSize}{0.3}

\begin{figure}[ht!]
	\centering
	\scriptsize
	\subfloat[UCMerced Land-use dataset] {
		\includegraphics[width=\convergenceFigSize\textwidth, keepaspectratio=true]{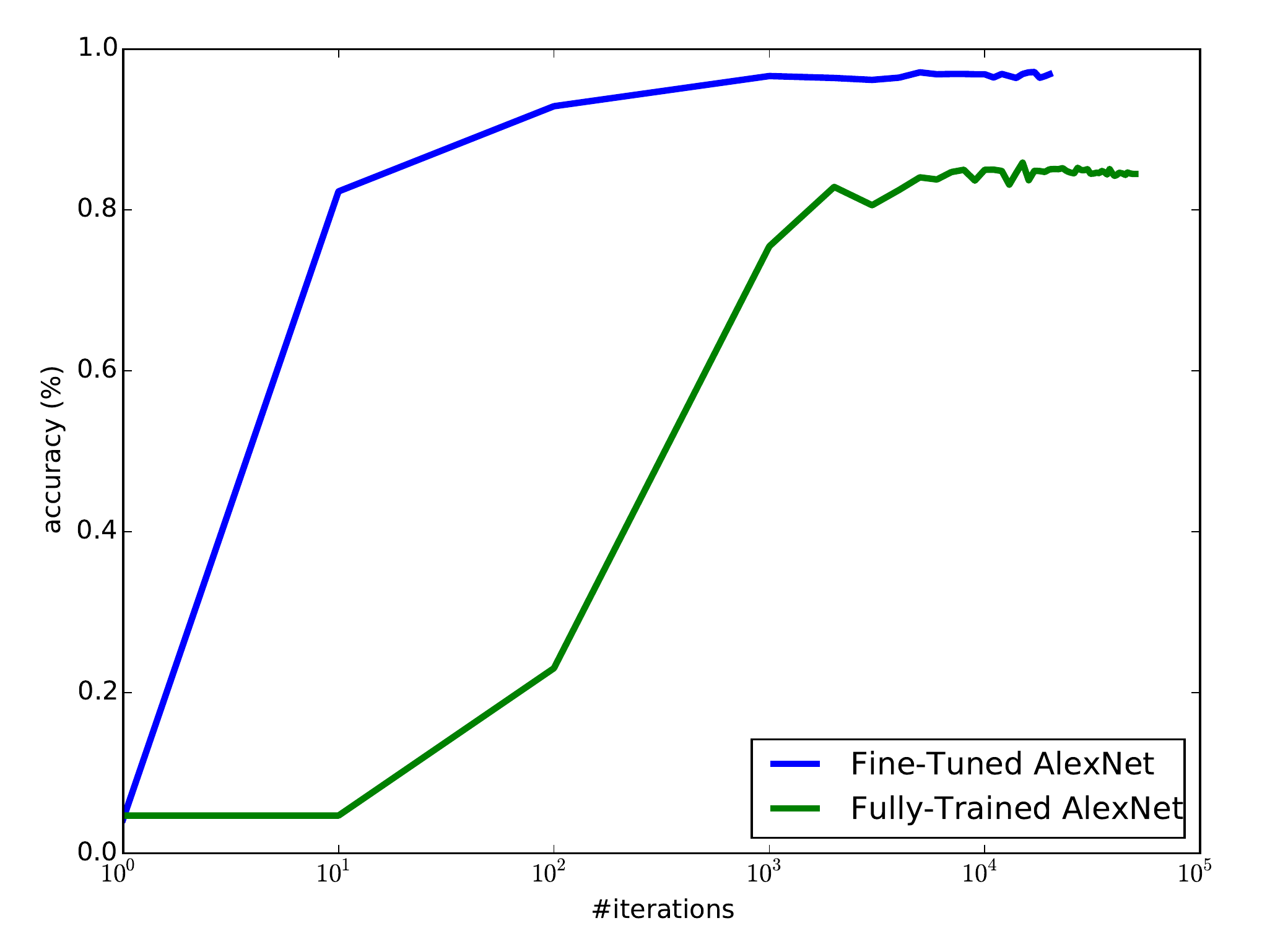}
	}
	\hspace{1mm}
	\subfloat[RS19 dataset]{
		\includegraphics[width=\convergenceFigSize\textwidth, keepaspectratio=true]{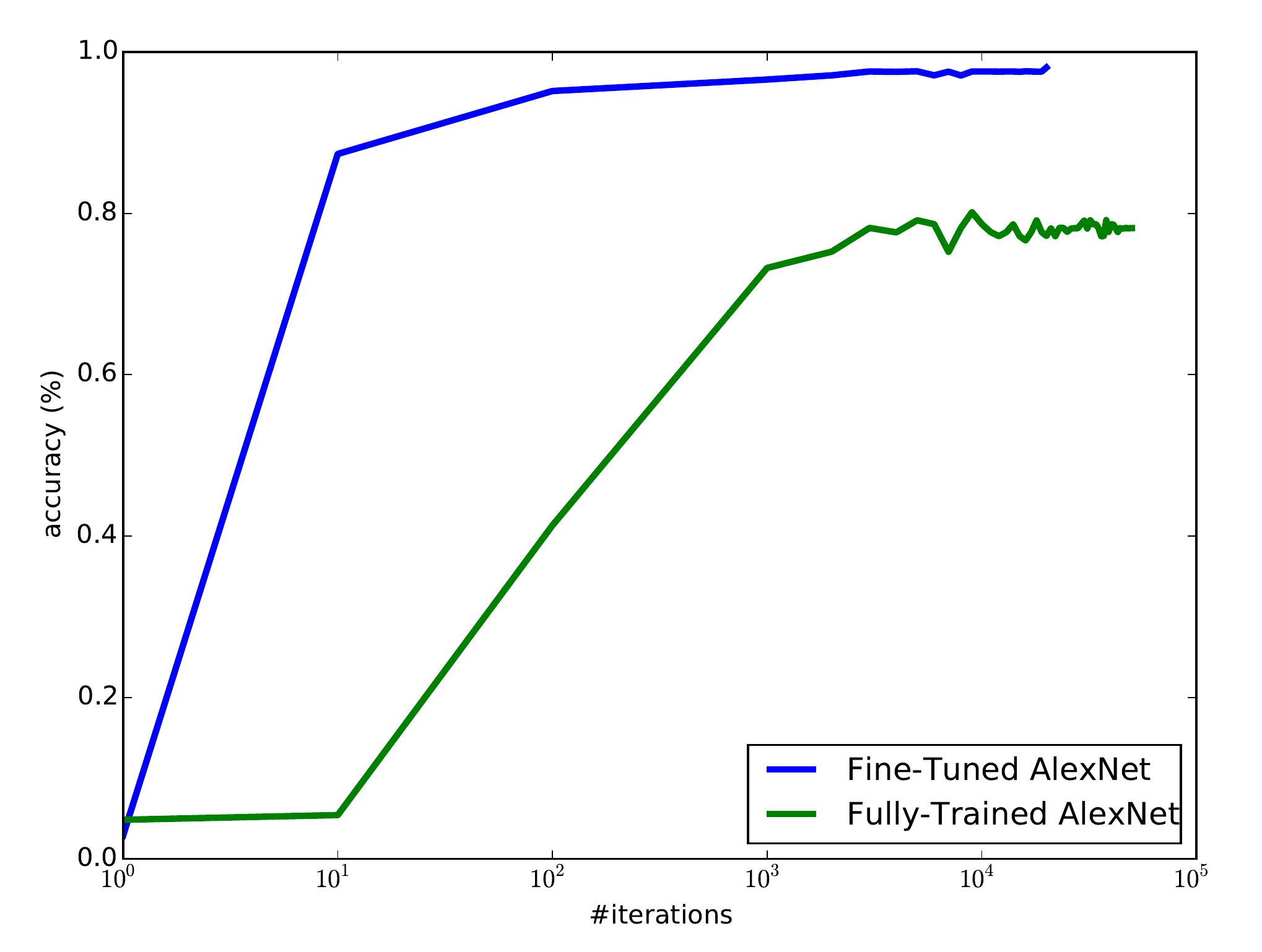}
	}
	\hspace{1mm}
	\subfloat[Brazilian Coffee Scenes dataset]{
		\includegraphics[width=\convergenceFigSize\textwidth, keepaspectratio=true]{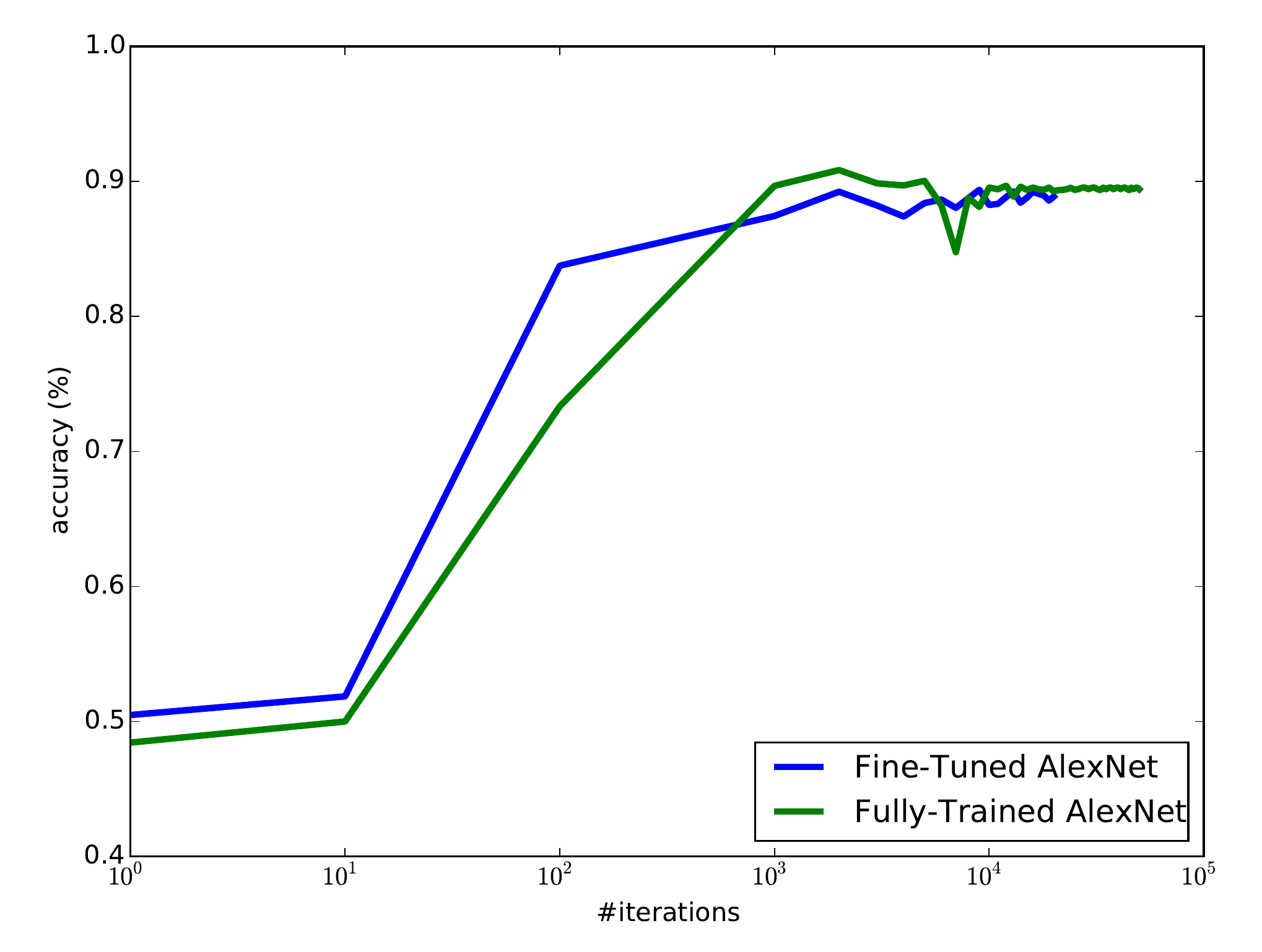}
	}
	\caption{Examples of converge of AlexNet for all datasets, considering Fold 1.}
	\label{fig:convergence}
\end{figure}

Other aspect to highlight is that full training (orange bars) was not a good strategy for datasets UCMerced and RS19. 
For RS19 dataset specially, there was a drop in accuracy in relation even to the original feature extractors (green bars), which were trained on ImageNet.
However, for Coffee Scenes dataset, the full-training strategy improved results in comparison with using the original ConvNets as feature extractors.

Another aspect of the results is that replacing the last softmax layer by SVM of almost every ConvNet was a better solution. 
The Fine-Tuned Descriptors (blue bars) and Fully-Trained Descriptors (orange bars) were usually superior than their counterparts with the softmax layer (red and pink bars, respectively).
In the Coffee Scenes, however, this difference was smaller.

Comparing the different ConvNets, we can see that their results are very similar in all the datasets.
GoogLeNet seems to be less affected by the full training process, as their results decreased less than the other ConvNets when comparing the Fine-Tuned and the Full-Trained versions. 
One possible reason is that GoogLeNet has less parameters to be learned and, as the datasets used are very small considering the requirements of the full training process, GoogLeNet was less affected.

Comparing the results of the ConvNets as feature extractors (green bars) in relation to fine tuning (red and blue bars), we can see that the original feature extractors, trained on ImageNet, are not too worse than the fine-tuned version of the ConvNets, specially for the UCMerced and RS19 datasets. 
The reason is that in such datasets, the edges and local structures of the images are more similar to everyday objects than in the Coffee Scenes dataset, in which the difference was higher in favor of fine-tuned ConvNets.
In the Coffee Scenes dataset, the textures and local structures are very different than everyday objects.

Comparing the results among the three datasets, we can see that the Coffee Scenes dataset has a different behavior for the ConvNets.
Full-trained networks achieve better accuracy in this datasets than in the others.
This maybe be motivated by the huge difference between the datasets, since UCMerced and RS19 datasets are aerial ones while Coffee scenes is a multi-spectral one.

As summary, we can recommend \textit{fine-tuning} as the strategy that tends to be more promising in different situations.
In addition, fine tuning is less costly than full training, which can represent another advantage when efficiency is a constraint.
On top of that, we can also recommend the use of the features extracted from the last layer of the fine-tuned network and then using SVM for the classification task, instead of using the softmax layer.




As shown in the experimental results, the best ConvNet configurations classify almost 100\% of the aerial imagens (UCMerced and RS19). 
Notwithstanding, the wrong classified images are really difficult, as can be noted in the examples shown in Figure~\ref{fig:wrong_predictions}. Notice how these misclassified samples are quite similar visually.

\newcommand{\wrongFigSize}{0.12}

\begin{figure}[ht!]
	\centering
	\scriptsize
	\subfloat[Freeway ${\xrightarrow[into]{missclassified}}$ Runaway] {
		\includegraphics[width=\wrongFigSize\textwidth, keepaspectratio=true]{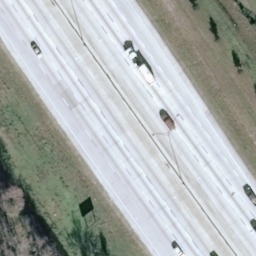}
		\includegraphics[width=\wrongFigSize\textwidth, keepaspectratio=true]{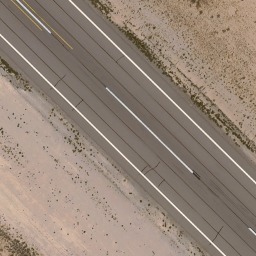}
	}
	\hspace{1mm}
	\subfloat[Medium Rensidential ${\xrightarrow[into]{missclassified}}$ Dense Rensidential]{
		\includegraphics[width=\wrongFigSize\textwidth, keepaspectratio=true]{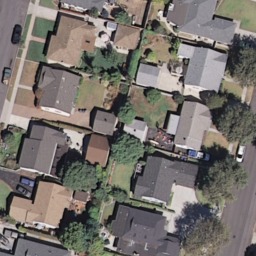}
		\includegraphics[width=\wrongFigSize\textwidth, keepaspectratio=true]{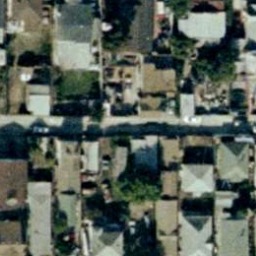}
	}
	\hspace{1mm}
	\subfloat[Dense Rensidential ${\xrightarrow[into]{missclassified}}$ Mobile Homepark]{
		\includegraphics[width=\wrongFigSize\textwidth, keepaspectratio=true]{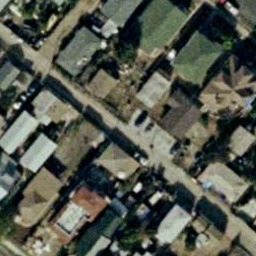}
		\includegraphics[width=\wrongFigSize\textwidth, keepaspectratio=true]{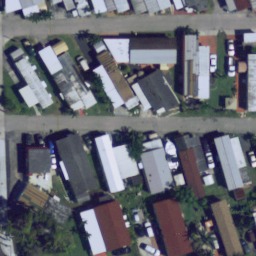}
	}
	\hspace{1mm}
	\subfloat[Commercial ${\xrightarrow[into]{missclassified}}$ Park]{
		\includegraphics[width=\wrongFigSize\textwidth, keepaspectratio=true]{commercial_30.jpg}
		\includegraphics[width=\wrongFigSize\textwidth, keepaspectratio=true]{Park_02.jpg}
	}
	\hspace{1mm}
	\subfloat[Farmland ${\xrightarrow[into]{missclassified}}$ Meadow]{
		\includegraphics[width=\wrongFigSize\textwidth, keepaspectratio=true]{Farmland-11.jpg}
		\includegraphics[width=\wrongFigSize\textwidth, keepaspectratio=true]{meadow_45.jpg}
	}
	\hspace{1mm}
	\subfloat[Forest ${\xrightarrow[into]{missclassified}}$ Moutain]{
		\includegraphics[width=\wrongFigSize\textwidth, keepaspectratio=true]{forest_09.jpg}
		\includegraphics[width=\wrongFigSize\textwidth, keepaspectratio=true]{Mountain_03.jpg}
	}
	\caption{Three examples of wrong predictions of each aerial dataset, UCMerced and RS19, for a fine-tuned AlexNet.
		(a)-(f) The first image is the misclassified one, while the second is a sample of the predicted class.
		Notice the similarity between the classes.}
	\label{fig:wrong_predictions}
\end{figure}

\subsection{Comparison with Baselines} \label{subsec:stateComparison}  

\begin{figure}[ht!]
	\centering
	\includegraphics[trim = 0mm 2mm 0mm 28mm, clip, width=\resultsFigSize\textwidth]{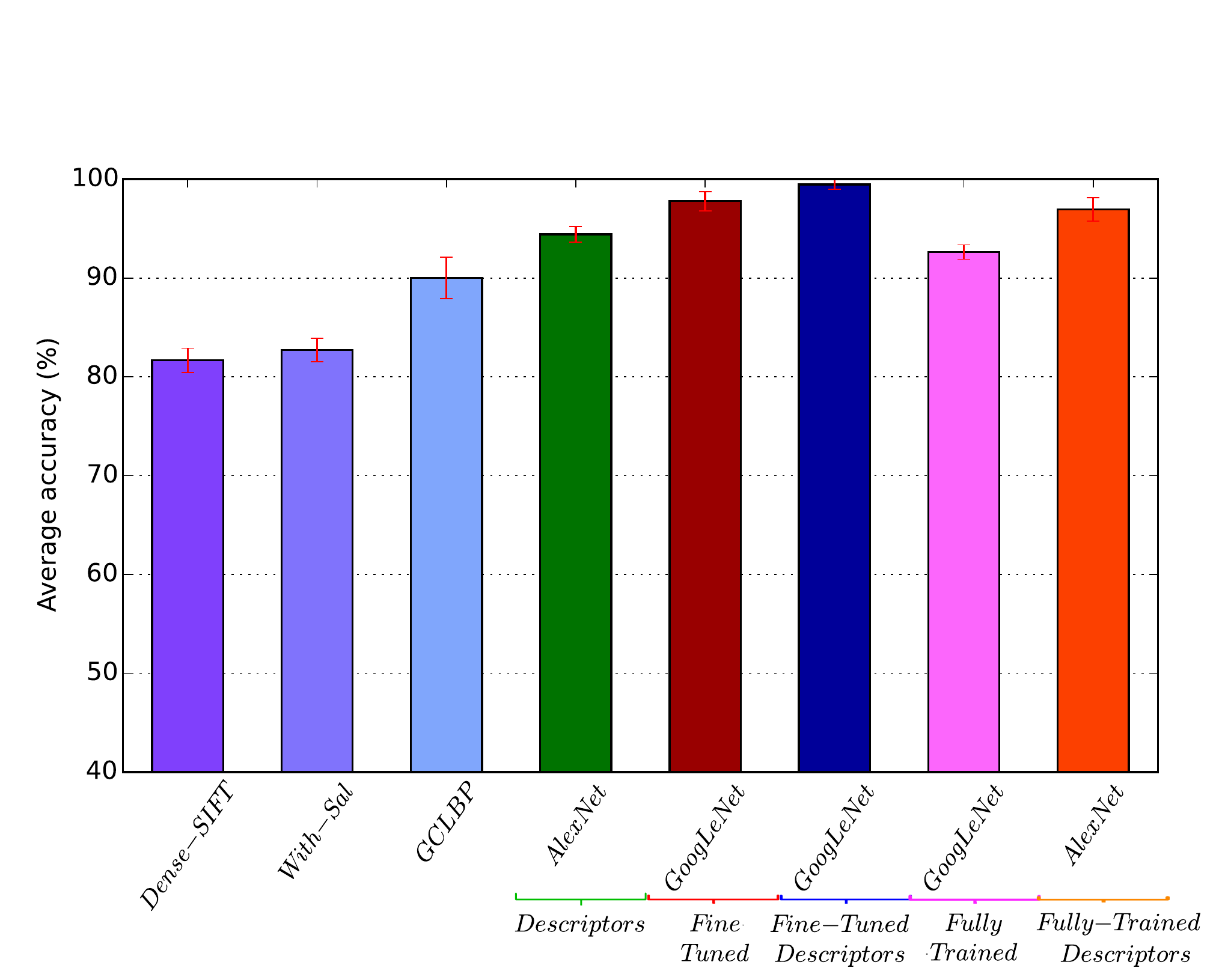}
	\caption{Comparison between state-of-the-art baselines and the best results of each strategy to exploit ConvNets for the UCMerced Land-use Dataset. The Fine-Tuned Descriptors extracted by GoogLeNet achieved the highest accuracy rates.}
	\label{fig:ucmerced_stateofart}
\end{figure}

\begin{figure}[ht!]
	\centering
	\includegraphics[trim = 0mm 2mm 0mm 28mm, clip, width=\resultsFigSize\textwidth]{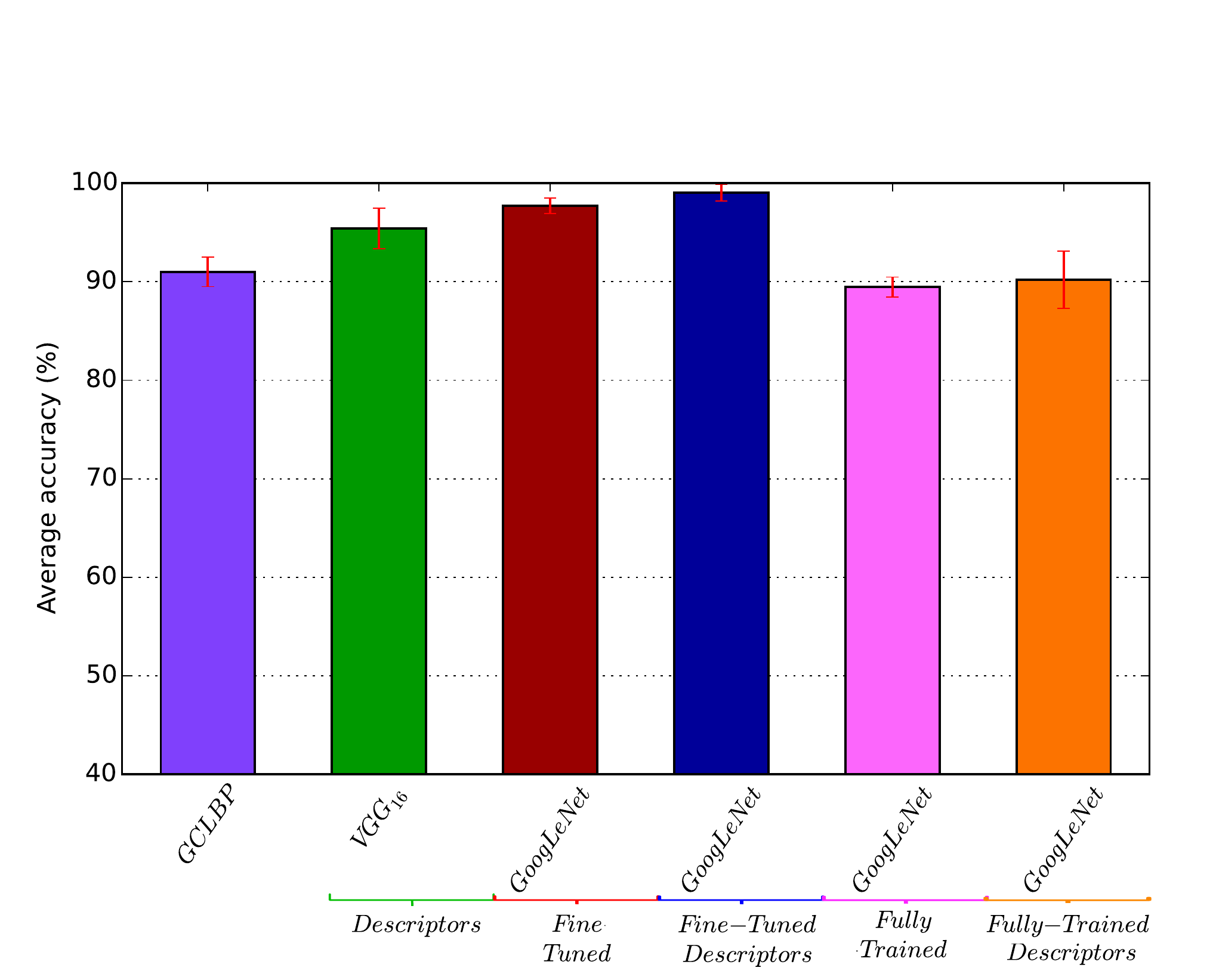}
	\caption{Comparison between state-of-the-art baselines and the best results of each strategy to exploit ConvNets for the RS19 Dataset. The Fine-Tuned Descriptors extracted by GoogLeNet achieved the highest accuracy rates.}
	\label{fig:rs19_stateofart}
\end{figure}

\begin{figure}[ht!]
	\centering
	\includegraphics[trim = 0mm 2mm 0mm 28mm, clip, width=\resultsFigSize\textwidth]{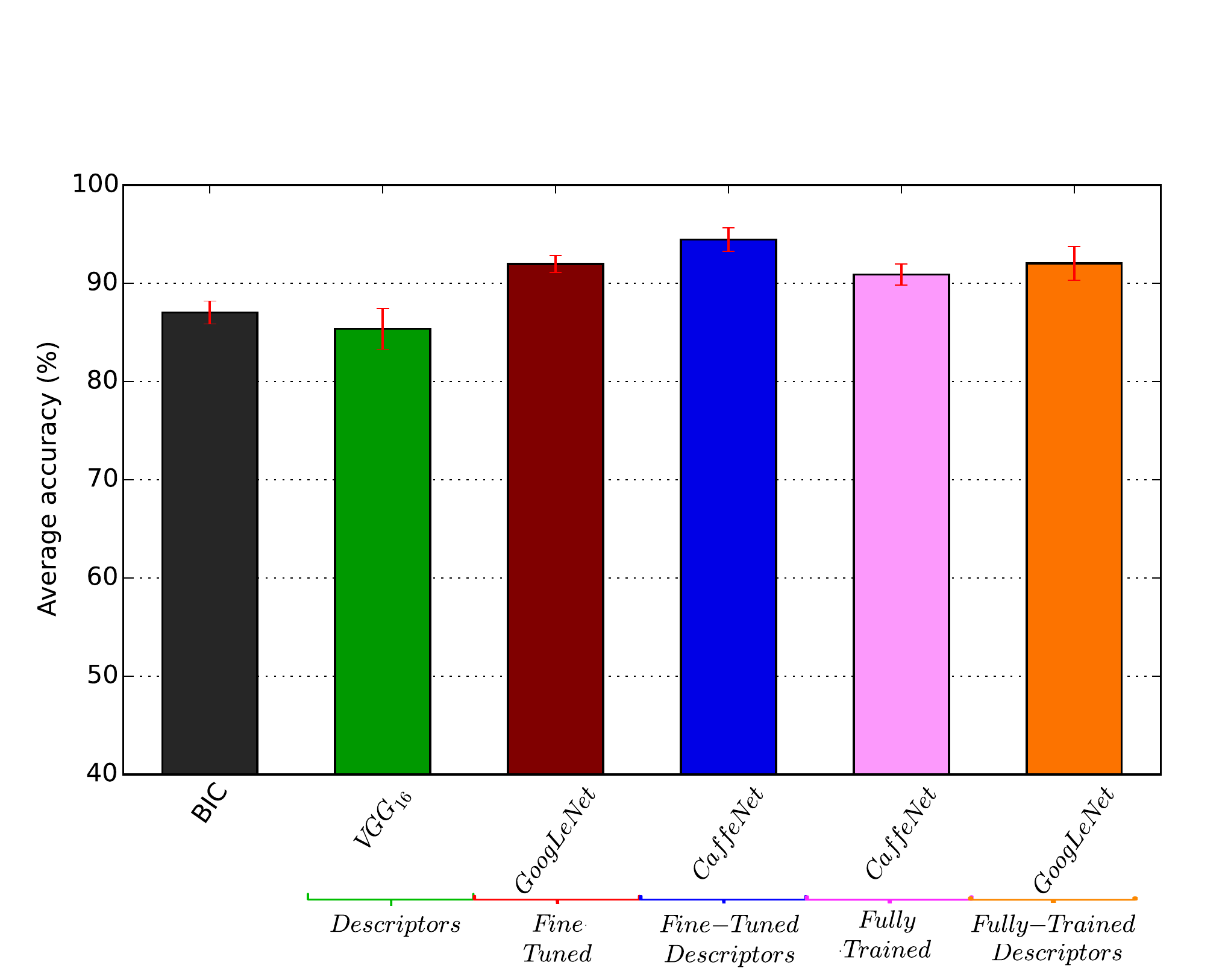}
	\caption{Comparison between state-of-the-art baselines and the best results of each strategy to exploit ConvNets for the Brazilian Coffee Scenes Dataset. The Fine-Tuned Descriptors extracted by CaffeNet achieved the highest accuracy rates.}
	\label{fig:coffee_stateofart}
\end{figure}

In this section, we compare the performance of the best results of each strategy for exploiting the existing ConvNets and state-of-the-art baselines.
Figures~\ref{fig:ucmerced_stateofart} to~\ref{fig:coffee_stateofart} show the comparison in terms of average accuracy.
As in previous section, the suffix ``Descriptors'' specify when ConvNets were used as feature extractor, being the deep features classified with linear SVM.

For the UCMerced dataset, we select three state-of-the-art baselines: (i) GCLBP~\cite{chen2015gabor}, (ii) With-Sal~\cite{Zhang2015:TGRS}, and (iii) Dense-Sift~\cite{cheriyadat2014unsupervised}.
The results presented in Figure~\ref{fig:ucmerced_stateofart} show that the baselines were outperformed by all strategies, except the full training one.
Furthermore, the classification of deep features extracted from the fine-tuned GoogLeNet with linear SVM achieve the best result of all (99.47$\pm$0.50), being closely followed by the fine-tuned GoogLeNet, which yielded 97.78$\pm$0.97, in terms of average accuracy.

For the RS19 dataset, we compare the best results of each strategy to the GCLBP~\cite{chen2015gabor} approach, which yielded 91.0$\pm$1.5.
The results presented in Figure~\ref{fig:rs19_stateofart} confirm the results obtained in the UCMerced dataset, since these two datasets are very similar: using linear SVM to classify deep features extracted from a fine-tuned GoogLeNet yielded the best result.
However, different from the previous dataset, the fully trained network did not outperform the baseline, being statistically similar.

For the Brazilian Coffee Scenes dataset, the only state-of-the-art result available is the one which was released with the dataset in our previous work~\cite{penattideep}, using the BIC descriptor, that we also present here in Section~\ref{subsec:generalization}.
Now, the best result for this dataset and current state-of-the-art is achieved by extracting deep features from the fine-tuned CaffeNet (94.45 $\pm$	1.20), as presented in Figure~\ref{fig:coffee_stateofart}.
Note that although BIC outperforms the ConvNet used as a descriptor, it is not true for the Full-Trained and Fine-Tuned.
It means that we can adjust the domain by using a full trained ConvNet. However, by using parameters obtained in other domain is useful to yield even better results.

\section{Conclusions} \label{sec:conclusions}

In this paper, we evaluated three strategies for exploiting existing ConvNets in different scenarios from the ones they were trained.
The objective was to understand the best way to obtain the most benefits from these state-of-the-art deep learning approaches in problems that usually are not suitable for the design and creation of new ConvNets from scratch.
Such scenarios reflect many existing applications, in which there is few labeled data.
We performed experiments evaluating the following strategies for exploiting the ConvNets: full training, fine tuning, and using as feature extractors.
The experiments considered six popular ConvNets (OverFeat networks~\cite{OverFeatIntegrated2014}, AlexNet~\cite{krizhevsky2012imagenet}, CaffeNet~\cite{jia2014caffe}, GoogLeNet~\cite{szegedy2014going}, $VGG_16$~\cite{simonyan2014very}, and PatreoNet~\cite{nogueira2015improving}) in three remote sensing datasets.

The results point that~\textit{fine tuning} tends to be the best strategy in different situations.
Specially, using the features of the fine-tuned network with an external classifier, linear SVM in our case, provides the best results.

As additional contributions of this work, we can point the evaluation of different ConvNets in each strategy mentioned in three remote sensing datasets, comparing their results with traditional low- and mid-level descriptors, as well as with state-of-the-art baselines of each dataset.
We can also understand the generalization power of the existing ConvNets when used as feature descriptors.
And finally, we obtained state-of-the-art results in the three datasets used (UCMerced land use, RS19, and Brazilian Coffee Scenes).

The presented conclusions open new opportunities towards a better spectral-spatial feature representation, which is still needed for remote sensing applications, such as agriculture or environmental monitoring. We also believe that these conclusions also applies to other domains, however, we would like to perform such evaluation as future work.
Another interesting opportunity for future work is to analyze the relation between the number of classes in the dataset, the number of parameters in the ConvNet, and their impact in the discrepancy between fine tuning and full training processes.

\section*{Acknowledgments}

This work was partially financed by CNPq (grant 449638/2014-6), CAPES, and Fapemig (APQ-00768-14).
The authors gratefully acknowledge the support of NVIDIA Corporation with the donation
of the GeForce GTX TITAN X GPU used for this research.

\bibliographystyle{elsarticle-num}
\bibliography{bibliography}

\end{document}